%% file: main.tex
\documentclass[journal]{depd/IEEEtran}

\usepackage{times}
\usepackage{epsfig}
\usepackage{graphicx}
\usepackage{amsmath}
\usepackage{amssymb}
\usepackage{amsthm}
\usepackage{bm}
\usepackage{subfig}
\usepackage{capt-of}
\usepackage{booktabs}
\usepackage[table]{xcolor}
\usepackage{mathtools}
\usepackage{siunitx}
\usepackage{stfloats}
\usepackage{arydshln}
\usepackage{longtable}
\usepackage[table]{xcolor}
\usepackage{textpos}

\usepackage{hyperref}

\newcommand{\etal}{\textit{et al}. }
\newcommand{\ie}{\textit{i}.\textit{e}., }
\newcommand{\eg}{\textit{e}.\textit{g}. }

\ifCLASSOPTIONcompsoc
  \usepackage[nocompress]{cite}
\else
  \usepackage{cite}
\fi

\begin{document}
	

\title{$\mathcal{G}$-softmax: Improving Intra-class Compactness and Inter-class Separability of Features}

\author{Yan~Luo,
        Yongkang~Wong,~\IEEEmembership{Member,~IEEE,}
        Mohan~Kankanhalli,~\IEEEmembership{Fellow,~IEEE,}
        and~Qi~Zhao,~\IEEEmembership{Member,~IEEE,}
\IEEEcompsocitemizethanks
			{
			\IEEEcompsocthanksitem Y. Luo and Q. Zhao are with the Department of Computer Science and Engineering, University of Minnesota, Minneapolis, MN, 55455.\protect\\
			E-mail: luoxx648@umn.edu, qzhao@cs.umn.edu
			\IEEEcompsocthanksitem Y. Wong and M. Kankanhalli are with the School of Computing, National University of Singapore, Singapore, 117417.
			E-mail: yongkang.wong@nus.edu.sg, mohan@comp.nus.edu.sg
			\IEEEcompsocthanksitem Qi Zhao is the corresponding author.
			}
}


\IEEEtitleabstractindextext{%
\input{depd/abstract.tex}

\begin{IEEEkeywords}
Deep learning, Multi-Label classification, Gaussian-based softmax, Compactness and separability.
\end{IEEEkeywords}}

\maketitle

\IEEEdisplaynontitleabstractindextext

\IEEEpeerreviewmaketitle

\begin{textblock*}{10cm}(-1.5cm,-15.8cm)
	{\footnotesize This is a pre-print and the published version is already available in TNNLS}
\end{textblock*}


\input{depd/intro.tex}
\input{depd/related.tex}
\input{depd/method.tex}
\input{depd/experiment.tex}
\input{depd/analysis.tex}
\input{depd/conclusion.tex}

\ifCLASSOPTIONcompsoc
 \section*{Acknowledgments}
\else
 \section*{Acknowledgment}
\fi
This research was funded in part by the NSF under Grant 1849107, in part by the University of Minnesota Department of Computer Science and Engineering Start-up Fund (QZ), and in part by the National Research Foundation, Prime Minister's Office, Singapore under its Strategic Capability Research Centres Funding Initiative.

\ifCLASSOPTIONcaptionsoff
  \newpage
\fi

\input{main.bbl}

\begin{IEEEbiography}[{\includegraphics[width=1in,height=1.25in,clip,keepaspectratio]{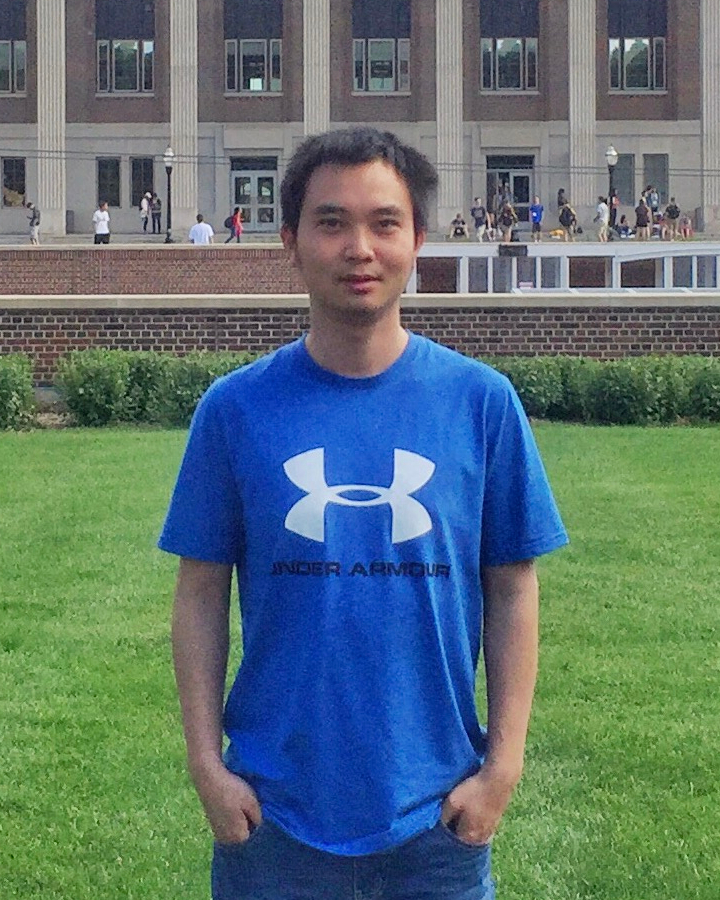}}]{Yan Luo}
received the B.Sc. degree in Computer Science from Xi'an, China, in 2008. He is currently pursuing the Ph.D. degree with the Department of Computer Science and Engineering, University of Minnesota at Twin Cities, Minneapolis, MN, USA. In 2013, he joined the Sensor-enhanced Social Media (SeSaMe) Centre at Interactive and Digital Media Institute, National University of Singapore, as a research assistant. In 2015, he joined the Visual Information Processing Laboratory at the National University of Singapore as a Ph.D. student. In 2017, he moved to University of Minnesota, Twin Cities, to continue his doctoral program. His research interests are in the areas of computer vision, computational visual cognition, and deep learning.
\end{IEEEbiography}


\begin{IEEEbiography}[{\includegraphics[width=1in,height=1.25in,clip,keepaspectratio]{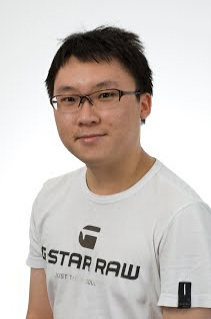}}]{Yongkang Wong}
is a senior research fellow at the School of Computing, National University of Singapore. He is also the Assistant Director of the NUS Centre for Research in Privacy Technologies (N-CRiPT). He obtained his BEng from the University of Adelaide and PhD from the University of Queensland. He has worked as a graduate researcher at NICTA's Queensland laboratory, Brisbane, OLD, Australia, from 2008 to 2012. His current research interests are in the areas of Image/Video Processing, Machine Learning, and Social Scene Analysis. He is a member of the IEEE since 2009.
\end{IEEEbiography}

\begin{IEEEbiography}[{\includegraphics[width=1in,height=1.25in,clip,keepaspectratio]{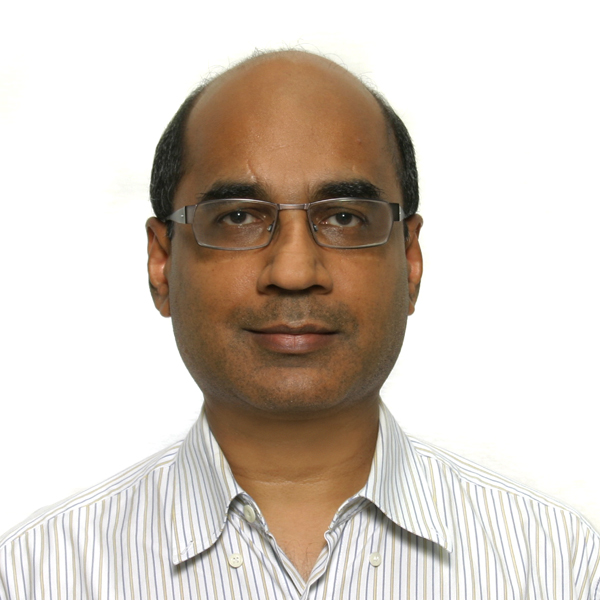}}]{Mohan~Kankanhalli}
is the Provost's Chair Professor at the Department of Computer Science of the National University of Singapore. He is the director with the N-CRiPT and also the Dean, School of Computing at NUS. Mohan obtained his BTech from IIT Kharagpur and MS \& PhD from the Rensselaer Polytechnic Institute. His current research interests are in Multimedia Computing, Multimedia Security, Image/Video Processing and Social Media Analysis. He is active in the Multimedia Research Community and is on the editorial boards of several journals. Mohan is a Fellow of IEEE since 2014.
%
\end{IEEEbiography}

\begin{IEEEbiography}[{\includegraphics[width=1in,height=1.25in,clip,keepaspectratio]{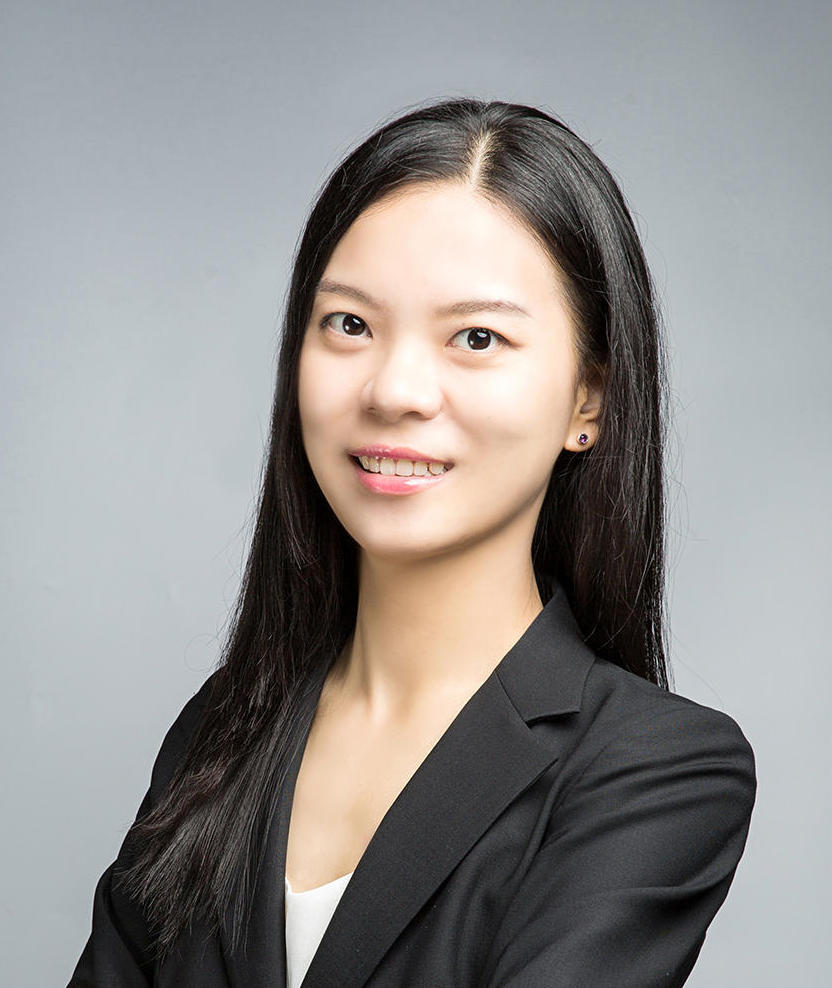}}]{Qi Zhao}
is an assistant professor in the Department of Computer Science and Engineering at the University of Minnesota, Twin Cities. Her main research interests include computer vision, machine learning, cognitive neuroscience, and mental disorders. She received her Ph.D. in computer engineering from the University of California, Santa Cruz in 2009. She was a postdoctoral researcher in the Computation \& Neural Systems, and Division of Biology at the California Institute of Technology from 2009 to 2011. Prior to joining the University of Minnesota, Qi was an assistant professor in the Department of Electrical and Computer Engineering and the Department of Ophthalmology at the National University of Singapore. She has published more than 50 journal and conference papers in top computer vision, machine learning, and cognitive neuroscience venues, and edited a book with Springer, titled Computational and Cognitive Neuroscience of Vision, that provides a systematic and comprehensive overview of vision from various perspectives, ranging from neuroscience to cognition, and from computational principles to engineering developments. She is a member of the IEEE since 2004.
\end{IEEEbiography}


\end{document}

%% file: depd/abstract.tex
\begin{abstract}
Intra-class compactness and inter-class separability are crucial indicators to measure the effectiveness of a model to produce discriminative features, where intra-class compactness indicates how close the features with the same label are to each other and inter-class separability indicates how far away the features with different labels are. In this work, we investigate intra-class compactness and inter-class separability of features learned by convolutional networks and propose a Gaussian-based softmax ($\mathcal{G}$-softmax) function that can effectively improve intra-class compactness and inter-class separability.  The proposed function is simple to implement and can easily replace the softmax function. We evaluate the proposed $\mathcal{G}$-softmax function on classification datasets (\ie CIFAR-10, CIFAR-100, and Tiny ImageNet) and on multi-label classification datasets (\ie MS COCO and NUS-WIDE). The experimental results show that the proposed $\mathcal{G}$-softmax function improves the state-of-the-art models across all evaluated datasets. In addition, analysis of the intra-class compactness and inter-class separability demonstrates the advantages of the proposed function over the softmax function, which is consistent with the performance improvement.
More importantly, we observe that high intra-class compactness and inter-class separability are linearly correlated to average precision on MS COCO and NUS-WIDE. This implies that improvement of intra-class compactness and inter-class separability would lead to improvement of average precision.

\end{abstract}

%% file: depd/intro.tex
\section{Introduction}
\label{sec:introduction}

\IEEEPARstart{M}{achine }learning is an important and fundamental component in visual understanding tasks. The core idea of supervised learning is to learn a model that explores the causal relationship between the dependent variables and the predictor variables. To quantify this relationship, the conventional approach is to make a hypothesis on the model, and feed the observed pairs of dependent variables and predictor parameters to the model for predicting future cases. For most learning problems, it is infeasible to make a perfect hypothesis that matches the underlying pattern, whereas a badly designed hypothesis often leads to a model that is more complicated than necessary and violates the principle of parsimony. Therefore, when designing or evaluating a model, the core objective is to seek a balance between two conflicting goals: how complicated a model should be to achieve accurate predictions, and how to design a model as simple as possible, but not simpler.

In the past decade, deep learning methods have significantly accelerated the development of machine learning research, where Convolutional Network (ConvNet) has achieved superior performance in numerous real-world visual understanding tasks~\cite{Girshick_CVPR_2014,Krizhevsky_NIPS_2012,Noh_ICCV_2015,Fu_PAMI_2017,Christian_RAS_2016,Hong_CVPR_2017,Zhou_ICCV_2017, Xu_CVPR_2017, Change_PAMI_2017,Hou_TNNLS_2015,Yuan_TNNLS_2015,Shao_TNNLS_2014}. 
Although their architectures vary with each other, the softmax function is widely used along with the cross entropy loss at the training phase~\cite{He_CVPR_2016,Krizhevsky_NIPS_2012,Lecun_IEEE_1998,Simonyan_ICML_2015,Szegedy_CVPR_2015}. 
The softmax function may not take the distribution pattern of previously observed samples into account to boost classification accuracy. In this work, we design a statistically driven extension of the softmax function that fits into the Stochastic Gradient Descent (SGD) scheme for end-to-end learning.
Furthermore, the final layer of the softmax function directly connects to the predictions and can maximally preserve generality for various ConvNets, \ie avoid complex modification of existing network architectures. 

\begin{figure}[!t]
	\centering
	\includegraphics[width=0.5\textwidth]{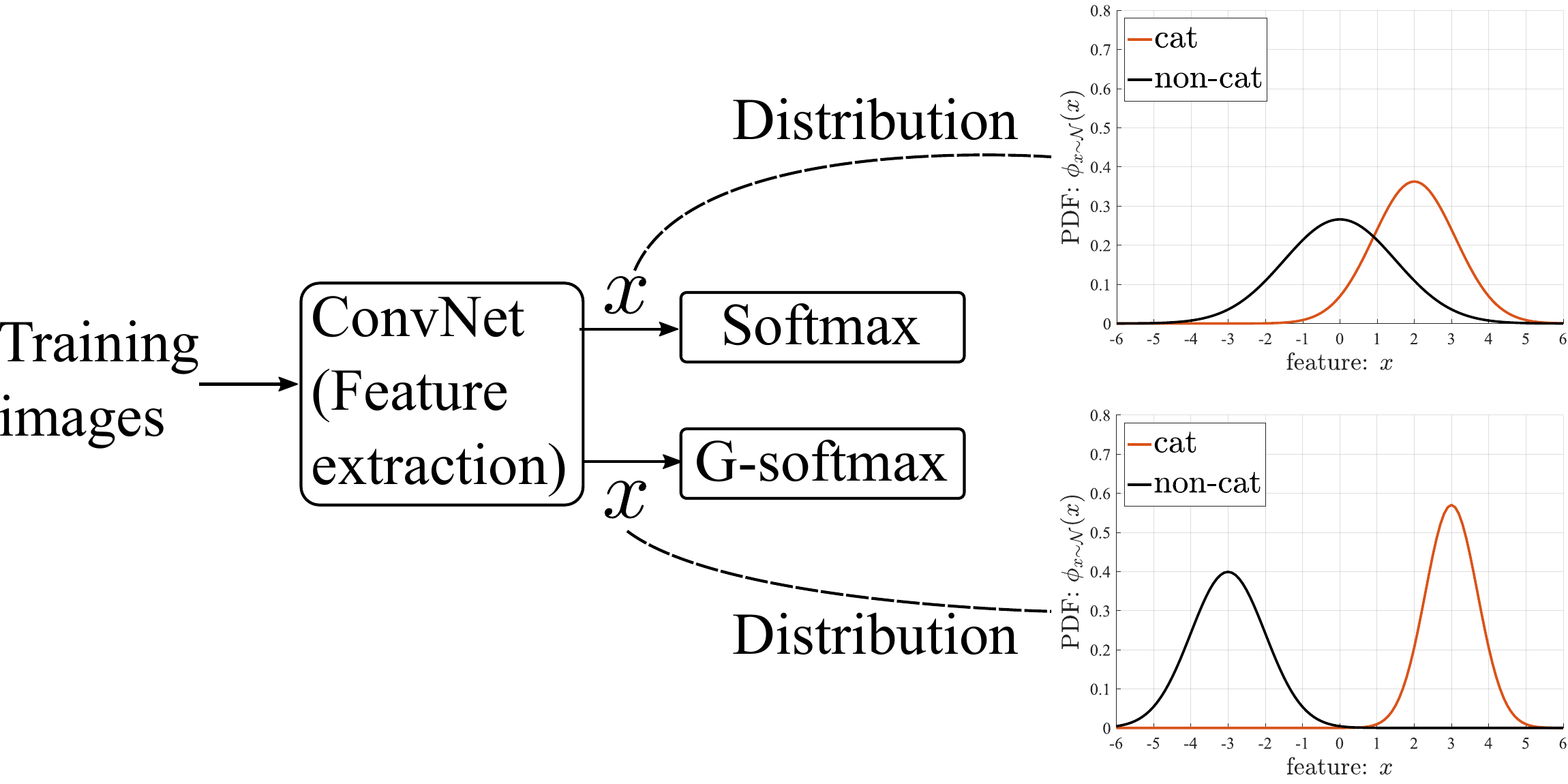}
	\caption
	{
		\small
		An illustration to show the benefits of improving inter-class separability and intra-class compactness. Given a model, an input image will be encoded to yield a discriminative feature $x$ that is used to compute the class-dependent confidence $p$. As shown in the figure, inter-class separability encourages to the distribution w.r.t. a label to be distant from the distributions w.r.t. the other labels, and intra-class compactness encourages the features w.r.t. the ground truth label to be close to the mean. PDF stands for probability density function.
	}
	\label{fig:teaser}
\end{figure}


\begin{figure*}[!t]
	\centering
	\includegraphics[width=1.0\textwidth]{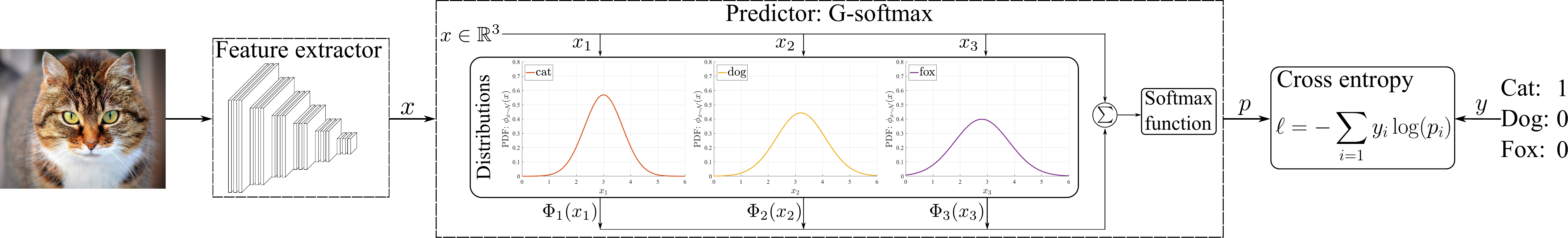}
	\caption
		{
		\small
		An illustration of the learning framework with the proposed $\mathcal{G}$-softmax function. The proposed $\mathcal{G}$-softmax function plays as a predictor to yield prediction confidences $p$ w.r.t. each class by taking input feature $x$ and its distribution into account. $\Phi_{i}(\cdot)$ is cumulative distribution function. The distributions are updated by gradient descent methods via back-propagation. In the mainstream ConvNets~\cite{He_CVPR_2016,Krizhevsky_NIPS_2012,Simonyan_ICML_2015,Szegedy_CVPR_2015}, the predictor is often the softmax function. 
		}
	\label{fig:illustrator}
\end{figure*}

Features are key for prediction in ConvNet Learning. According to the central limit theorem~\cite{Kallenberg_book_1997}, the arithmetic mean of a sufficiently large number of iterates of i.i.d. random variables, each with a finite expected value and variance, can be approximately normally distributed even if the original variables are not normally distributed. This makes the Gaussian distribution generally valid in a great variety of contexts. Following this line of thought, online learning methods \cite{Crammer_NIPS_2008,Dredze_ICML_2008,Wang_ICML_2012} assumed that the weights follow Gaussian distribution and make use of its distribution pattern for classification. Given a large-scale training data~\cite{Russakovsky_IJCV_2015}, the underlying distributions of discriminative features generated by ConvNets can be modeled. This distribution pattern has not been fully explored in existing literature.

Intra-class compactness and inter-class separability of features are generally correlated to the quality of the learned features. If intra-class compactness and inter-class separability are simultaneously maximized, the learned features are more discriminative \cite{Liu_ICML_2016}. We introduce a variant of the softmax function, named Gaussian-based softmax ($\mathcal{G}$-softmax) function, which aims to improve intra-class compactness and inter-class separability as shown in Figure \ref{fig:teaser}. We assume that features are distributed according to Gaussian distributions. Consequently, Gaussian cumulative distribution function (CDF) is used in prediction and normalization to generate the final confidence in a soft form.

Figure~\ref{fig:illustrator} demonstrates the role and position of the proposed $\mathcal{G}$-softmax function in a supervised learning framework. Given the training samples, the feature extractor would extract the features and then pass them to the predictor for inference. In this work, we follow the mainstream deep learning framework where the feature extractor is modeled with a ConvNet. The proposed $\mathcal{G}$-softmax function is able to replace the softmax function. 
The contributions can be summarized as: 
\begin{itemize}
	\item With the general assumption, \ie features w.r.t. a class are subject to a Gaussian distribution, we propose the $\mathcal{G}$-softmax function which models the distributions of features for better prediction. The experiments on CIFAR-10, CIFAR-100 \cite{Krizhevsky_Citeseer_2009} and Tiny ImageNet\footnote{\url{https://tiny-imagenet.herokuapp.com/}} show that the proposed $\mathcal{G}$-softmax function consistently outperforms the softmax and L-softmax function on various state-of-the-art models. Also, we apply the proposed $\mathcal{G}$-softmax function to solve the multi-label classification problem, which yields better performance than the softmax function on MS COCO \cite{Lin_ECCV_2014} and NUS-WIDE~\cite{Chua_CIVR_2009}. The source code is available\footnote{\url{https://gitlab.com/luoyan/gsoftmax}} and is easy for use.
	\item The proposed $\mathcal{G}$-softmax function can quantify the compactness and separability. Specifically, for each learned Gaussian distribution, the corresponding mean and variance indicate the center and compactness of the predictor. 
	\item 
	In our analysis of correlation between intra-class compactness (or inter-class separability) and average precision,  we observe that high intra-class compactness and inter-class separability are linearly correlated to average precision on MS COCO and NUS-WIDE. This implies that improvement of intra-class compactness and inter-class separability would leads to improvement of average precision.

\end{itemize}

%% file: depd/related.tex
\section{Related Works}
\label{sec:related}

\noindent\textbf{Gaussian-based Online Learning}.
We first review the Gaussian-based online learning methods. In the online learning context, the training data are provided in a sequential order to learn a predictor for unobserved data. These methods usually make some assumptions to minimize the cumulative disparity errors between the ground truth and predictions over the entire sequence of instances~\cite{Crammer_JMLR_2006,Crammer_NIPS_2008,Dredze_ICML_2008,Rosenblatt_Psychological_1958,Wang_ICML_2012}. In this sense, these works can give some guidance and inspiration for designing a flexible mapping function.

In contrast to Passive-Aggressive model \cite{Crammer_JMLR_2006}, Dredze \etal \cite{Dredze_ICML_2008} made an explicit assumption on the weights {\small $w \in \mathbb{R}^{m}$: $w \sim \mathcal{N}(\mu, \Sigma)$}, where $\mu$ is the mean of the weights $w$ and {\small $\Sigma \in \mathbb{R}^{m\times m}$} is a covariance matrix for the underlying Gaussian distribution. Given an input instance {\small $x_{i} \in \mathbb{R}^{m}$} with the corresponding label $y_{i}$, the multivariate Gaussian distribution over weight vectors induces a univariate Gaussian distribution over the margin: {\small $y_{i}(\langle w, x_{i}\rangle ) \sim \mathcal{N}(y_{i}(\langle \mu, x_{i} \rangle), x_{i}^{\top}\Sigma x_{i})$}, where $\langle\cdot, \cdot\rangle$ is the inner product operation. Hence, the probability of a correct prediction is {\small $Pr[y_{i}(\langle w, x_{i}\rangle) \ge 0]$}. The objective is to minimize the Kullback-Leibler divergence between the current distribution and the ideal distribution with the constraint that the probability of a correct prediction is not smaller than the confidence hyperparameter {\small $\beta \in [0,1]$}, \ie {\small $Pr[y_{i}(\langle w, x_{i}\rangle) \ge 0] \ge \beta$}. With the mean of the margin {\small $\mu_{M} = y_{i}(\langle\mu, x_{i} \rangle)$} and the variance {\small $\sigma_{M}^{2} = x_{i}^{\top}\Sigma x_{i}$}, the constraint can lead to {\small $y_{i}(\langle\mu, x_{i}\rangle) \ge \Phi^{-1}(\beta)\sqrt{x_{i}^{\top}\Sigma x_{i}}$}, where $\Phi$ is the cumulative function of the Gaussian distribution. This inequality is used as a constraint in optimization in practice. However, it is not convex with respect to $\Sigma$ and Dredze~\etal \cite{Dredze_ICML_2008} linearized it by omitting the square root: {\small $y_{i}(\langle\mu, x_{i}\rangle) \ge \Phi^{-1}(\beta)\left( x_{i}^{\top}\Sigma x_{i} \right)$}. To solve this non-convex problem, Crammer \etal \cite{Crammer_NIPS_2008} discovered that a change of variable helps to maintain the convexity, \ie when {\small $\Sigma = \Upsilon^{2}$}, the constraint becomes {\small $y_{i}(<\mu, x_{i}>) \ge \Phi^{-1}(\beta)\| \Upsilon x_{i} \|$}. The confidence weighted method \cite{Crammer_NIPS_2008} employs an aggressive updating strategy by changing the distribution to satisfy the constraint imposed by the current instance, which may incorrectly update the parameters of the distribution when handling a mislabeled instance. Therefore, Wang \etal \cite{Wang_ICML_2012} introduced  a trade-off parameter $C$ to balance the passiveness and aggressiveness.

The aforementioned online learning methods~\cite{Crammer_NIPS_2008,Dredze_ICML_2008,Wang_ICML_2012} hypothesize that the weights are subject to a multivariate Gaussian distribution and pre-define a confidence hyperparameter $\beta$ to formalize a constraint for optimization. Nevertheless, the weights are learned based on the training data, putting hypothesis on the weights could be similar to put the cart before the horse. Moreover, such confidence hyperparameter may not be flexible or adaptive for various datasets. In this work, we instead hypothesize that the features are subject to Gaussian distribution and there is no confidence hyperparameter. To update the weights, \cite{Crammer_NIPS_2008,Dredze_ICML_2008,Wang_ICML_2012} apply the Lagrangian method to compute the optimal weights. This mechanism does not straightforwardly fit into SGD scheme. Along the same line, this work is motivated to investigate how to incorporate the Gaussian assumption in SGD.

\noindent\textbf{Softmax Function in ConvNet Learning}.
The success of ConvNets is largely attributed to the layer-stacking mechanism. Despite its effectiveness in complex real-world visual classification, this mechanism will result in co-adaptation and overfitting. To prevent the co-adaptation problem, Hinton~\etal \cite{Hinton_arXiv_2012} proposed a method which randomly omits a portion of neurons in a feedforward network. Then, Srivastava~\etal \cite{Srivastava_JMLR_2014} introduced the dropout unit to minimize overfitting and presented a comprehensive investigation of its effect in ConvNets. Similar regularization methods are also proposed in \cite{Goodfellow_ICML_2013} and \cite{Wan_ICML_2013}. Instead of modifying the connection between layers, \cite{Zeiler_arXiv_2013} replaced the deterministic pooling with the stochastic pooling for regularizing ConvNets. The proposed $\mathcal{G}$-softmax function can be used together with these models to offer better general ability.
We posit a general assumption and establish Gaussian distributions over the feature space at the final layer, \ie the softmax module. In other words, the proposed $\mathcal{G}$-softmax function is general for most ConvNets without requiring much modification of the network structure. 

ConvNets~\cite{Huang_CVPR_2017,Zagoruyko_BMVC_2016,He_CVPR_2016,Lecun_IEEE_1998,Krizhevsky_NIPS_2012,Simonyan_ICML_2015,Szegedy_CVPR_2015, Yu_CVPR_2017} have strong representational ability in learning invariant features. Although their architectures vary with each other, the softmax function is widely used along with cross entropy loss at the training phase. Hence, the softmax module is important and general for ConvNets. Liu~\etal \cite{Liu_ICML_2016} introduced a large-margin softmax function to enhance the compactness and the separability from a geometric perspective. Substantially, the large-margin softmax function is fundamentally similar to the softmax function, \ie both use the exponential function, while having different inputs for the exponential function. In contrast, we model the mappings between features and ground truth labels as Gaussian CDF. Similar to the softmax function, we utilize normalization to identify the maximal element but not its exact value.

\noindent\textbf{Multi-label Classification}.
Multi-label classification is a special case of multi-output learning tasks. Read \etal \cite{Read_ECMLKDD_2009} proposed the classifier chain model to model label correlations. In particular, label order is important for chain classification models. A dynamic programming based classifier chain algorithm \cite{Liu_NIPS_2015} was proposed to find the globally optimal label order for the classifier chain models. Shen \etal \cite{Shen_NNLS_2018} introduced Co-Embedding and Co-Hashing  method that explores the label correlations from the perspective of cross-view learning to improve prediction accuracy and efficiency. On the other hand, the classifier chain model does not take the order of difficulty of the labels into account. Therefore, the easy-to-hard learning paradigm \cite{Liu_JMLR_2017b} was proposed to make good use of the predictions from simple labels to improve the predictions from hard labels. 
Liu \etal \cite{Liu_JMLR_2017a} presented comprehensively theoretical analysis on the curse of dimensionality of decision tree models and introduced a sparse coding tree framework for multi-label annotation problems. 
In multi-label prediction, a large margin metric learning paradigm \cite{Liu_AAAI_2015} was introduced to reduce the complexity of decoding procedure in canonical correlation analysis and maximum margin output coding methods.
Liu \etal \cite{Liu_PAMI_2018} introduced a large margin metric learning method to efficiently learn an appropriate distance metric for multi-output problems with theoretical guarantee.

Recently, there have been attempts to apply deep networks in multi-label classification, especially ConvNets and Recurrent Neural Networks (RNNs), for their promising performance in various vision tasks. In \cite{Wang_CVPR_2016}, ConvNet and RNN are utilized together to explicitly exploit the label dependencies. In contrast to \cite{Wang_CVPR_2016}, \cite{Zhang_arXiv_2016} proposed a regional latent semantic dependencies model to predict small-size objects and visual concepts by exploiting the label dependencies at the regional level. Similarly, \cite{Durand_CVPR_2016} automatically selected relevant image regions from global image labels using weakly supervised learning. Zhao \etal\cite{Zhao_BMVC_2016} reduced irrelevant and noisy regions with the help of region gating module. These region proposal based methods usually suffer from redundant computation and sub-optimal performance. Wang \etal\cite{Wang_2017_ICCV} addressed these problems by developing a recurrent memorized-attention module, and the module allows to locate attentional regions from the ConvNet's feature maps. Instead of utilizing the label dependencies, \cite{Li_CVPR_17} proposed a novel loss function for pairwise ranking, and the loss function is smooth everywhere so that it is easy to optimize within ConvNets. Also, there are two works that focus on improving the architectures of the networks for multi-label classification \cite{Zhu_CVPR_2017, Durand_CVPR_2017}. In this work, we adopt a common baseline, \ie ResNet-101 \cite{He_CVPR_2016}, which is widely used in the state-of-the-art models \cite{Zhu_CVPR_2017, Durand_CVPR_2017}.

%% file: depd/method.tex
\section{Methodology}
\label{sec:method}

\subsection{$\mathcal{G}$-Softmax Function}
\label{sec:sub1}

Logistic function, \ie sigmoid function, and hyperbolic tangent function are widely used in deep learning, whose graphs are ``S-shaped'' curves. Their curves imply a graceful balance between linearity and non-linearity~\cite{Menon_NN_1996}. The Gaussian CDF has the same monotonicity as logistic and hyperbolic tangent function and shares similar shapes. It makes the Gaussian CDF a potential substitute with the capability to model the distribution pattern with class dependent $\mu$ and $\sigma$. Fundamentally, softmax function in mainstream deep learning models is the normalized exponential function, which is a generalization of the logistic function. In this work, the proposed $\mathcal{G}$-softmax function uses the Gaussian CDF to substitute the exponential function.

Similar to the softmax loss, we use cross entropy as the loss function, \ie
\begin{align}
	\begin{split}
		\ell &=  -\sum_{i=1}^{m} y_{i} \log(p_{i}),
	\end{split}
	\label{eqn:loss}
\end{align}

\noindent
where $\ell$ is the loss, $y_{i} \in \{0, 1\}$ is the label with respect to the $i$-th category, $p_{i}$ is the prediction confidence with respect to the $i$-th category, and $m$ is the number of categories. Conventionally, given features $x$ that with respect to various labels, $p_{i}$ is given by the softmax function
\begin{align}
\begin{split}
p_{i} = \frac{e^{x_{i}}}{\sum_{j=1}e^{x_{j}}}.
\end{split}
\label{eqn:softmax}
\end{align}
The softmax function can be considered to represent a categorical distribution. By normalizing exponential function, the largest value is highlighted and the other values are suppressed significantly.
As discussed in Section~\ref{sec:related}, \cite{Crammer_NIPS_2008,Dredze_ICML_2008,Wang_ICML_2012} hypothesized that the classification margin is subject to a Gaussian distribution. Slightly differently, we assume that the deep features $x_{i}$ with respect to the $i$-th category is subject to a Gaussian distribution, \ie {\small $x_{i} \sim \mathcal{N}(\mu_{i}, \sigma_{i}^{2})$}. In this work, we define the proposed $\mathcal{G}$-softmax function as
\begin{align}
	\begin{split}
		p_{i} =  \frac{\exp\left(\overbrace{x_{i}}^{\text{activation}} + \overbrace{\lambda \Phi(x_{i}; \mu_{i}, \sigma_{i})}^{\text{distribution term}} \right)}{\sum_{j=1} \exp\left(x_{j} + \lambda \Phi(x_{j}; \mu_{j}, \sigma_{j}) \right)}.
	\end{split}
	\label{eqn:gsoftmax}
\end{align}


\noindent
where $\lambda$ is a parameter controlling the width of CDF along y-axis. We can see that if $\lambda=0$, Equation~(\ref{eqn:gsoftmax}) becomes the conventional softmax function. $\Phi$ is the CDF of a Gaussian distribution, that is 
\begin{align}
	\begin{split}
		\Phi(x_{i}; \mu_{i}, \sigma_{i}) = \frac{1}{2} \,	\text{erf}\left(-\frac{\sqrt{2} {\left(\mu_{i} - x_{i}\right)}}{2 \, \sigma_{i}}\right) +	\frac{1}{2}, \\
		\text{where}  \ \text{erf(}z\text{)} = \frac{1}{\sqrt{\pi}} \int_{-z}^{z}e^{-t^{2}}dt,
	\end{split}
\end{align}
\noindent
where $\mu$ and $\sigma$ are the mean and standard deviation, respectively. For simplicity, we denote {\small $\Phi(x_{i}; \mu_{i}, \sigma_{i})$} as {\small $\Phi_{i}$} in the following paragraphs.

Comparing to the softmax function (\ref{eqn:softmax}), the proposed $\mathcal{G}$-softmax function takes the feature distribution into account, \ie the distribution term in (\ref{eqn:gsoftmax}). This formulation leads to two advantages. First, it enables to approximate a large variety of the distributions w.r.t. every class on the training samples, whereas the softmax function only learns from current observing sample. Second, with distribution parameters $\mu$ and $\sigma$, it is straightforward to quantify intra-class compactness and inter-class separability. In other words, the proposed $\mathcal{G}$-softmax function is more analytical than the softmax function.

The proposed $\mathcal{G}$-softmax function can work with any ConvNets, such as VGG \cite{Simonyan_ICML_2015} and ResNet~\cite{He_CVPR_2016}.
In this work, we make $f(x_{l}) = \Phi(x_{l})$, and $l$ is not an arbitrary layer but the fully-connected layer. When $x_{l+1}=x_{l}+\lambda\Phi(x_{l})$, $x_{l+1}$ is prone to shift towards the positive axis direction because $\Phi(x_{l}) \in [0,1]$. 
The curve of $\Phi$ has a similar shape as that of logistic function and hyperbolic tangent function, and can accurately capture the distribution of $x$. As discussed in Section~\ref{sec:related}, the online learning methods \cite{Crammer_NIPS_2008,Dredze_ICML_2008,Wang_ICML_2012} considered the features as a Gaussian distribution and use Kullback-Leibler divergence (KLD) between the estimated distribution and the optimal distribution. Since their formulations involve the unknown optimal Gaussian distribution, they had to apply the Lagrangian to optimize and approximate $\mu$ and $\sigma$. This may not fit the back-propagation in modern ConvNets  which commonly use SGD as a solver. 

To optimize $\mu$, we have to compute the partial derivatives of Equation~(\ref{eqn:loss}) using the chain rule,
	\begin{align}
	\begin{split}
	\frac{\partial \ell}{\partial \mu_{i}}  &= \frac{\partial }{\partial \Phi_{i}} \left( -\sum_{i=1} y_{i} \log \frac{\exp(x_{i} + \lambda \Phi_{i})}{\sum_{j} \exp(x_{j} + \lambda\Phi_{j})} \right) \frac{\partial \Phi_{i}}{\partial \mu_{i}}\\
	&= \lambda\left( \left( \left(x_{i} + \lambda \Phi_{i}\right)\sum_{j}^{}y_{j} \right) - y_{i} \right) \frac{\partial \Phi_{i}}{\partial \mu_{i}}.
	\end{split}
	\label{eqn:lossderiv_mu}
	\end{align}

\noindent
Usually, $\sum_{j}^{}y_{j}$ equals to 1 due to the normalization. Similarly, we can obtain the partial derivatives with respect to $\sigma$,
	\begin{align}
	\begin{split}
	\frac{\partial \ell}{\partial \sigma_{i}}  = \lambda\left( \left( \left(x_{i} + \lambda \Phi_{i}\right)\sum_{j}^{}y_{j} \right) - y_{i} \right) \frac{\partial \Phi_{i}}{\partial \sigma_{i}}.
	\end{split}
	\label{eqn:lossderiv_sigma}
	\end{align}

\noindent
According to the CDF, \ie Equation~(\ref{eqn:gsoftmax}), the derivatives with respect to $\mu$ and $\sigma$ are
\begin{align}
\frac{\partial \Phi_{i}}{\partial \mu_{i}} &= -\frac{\sqrt{2} e^{\left(-\frac{{\left(\mu_{i} - x_{i}\right)}^{2}}{2 \, \sigma_{i}^{2}}\right)}}{2 \, \sqrt{\pi} \sigma_{i}}, \label{eqn:pder_mu}\\
\frac{\partial \Phi_{i}}{\partial \sigma_{i}} &= \frac{\sqrt{2} {\left(\mu_{i} - x_{i}\right)} e^{\left(-\frac{{\left(\mu_{i} - x_{i}\right)}^{2}}{2 \, \sigma_{i}^{2}}\right)}}{2 \,	\sqrt{\pi} \sigma_{i}^{2}}. \label{eqn:pder_sigma}
\end{align}

\noindent
Plugging (\ref{eqn:pder_mu}) and (\ref{eqn:pder_sigma}) into (\ref{eqn:lossderiv_mu}) and (\ref{eqn:lossderiv_sigma}), partial derivatives of $\mu$ and $\sigma$ are

\begin{small}
\begin{align}
\frac{\partial \ell}{\partial \mu_{i}}  &= \lambda\left(y_{i} - \left(x_{i} + \lambda \Phi_{i}\right)\sum_{j}^{}y_{j} \right) \!\!  \frac{\sqrt{2} e^{\left(-\frac{{\left(\mu_{i} - x_{i}\right)}^{2}}{2 \, \sigma_{i}^{2}}\right)}}{2 \, \sqrt{\pi} \sigma_{i}}, \label{eqn:lossder_final_mu}\\
\frac{\partial \ell}{\partial \sigma_{i}}  &= \lambda\left( \!\!\! \left( \! \left(x_{i} + \lambda \Phi_{i}\right)\sum_{j}^{}y_{j} \! \right) - y_{i} \! \right) \!\! \frac{\sqrt{2} {\left(\mu_{i} - x_{i}\right)} e^{\left(-\frac{{\left(\mu_{i} - x_{i}\right)}^{2}}{2 \, \sigma_{i}^{2}}\right)}}{2 \,	\sqrt{\pi} \sigma_{i}^{2}}.
\label{eqn:lossder_final_sigma}
\end{align}
\end{small}

In the back-propagation of ConvNets, the chain rule requires the derivatives of upper layers to compute the weight derivatives of lower layers. Therefore, $\frac{\partial \ell}{\partial x_{i}}$ is needed to pass backwards the lower layers. Because $\frac{\partial \ell}{\partial x_{i}}$ has the same form as $\frac{\partial \ell}{\partial \mu_{i}}$ in Equation (\ref{eqn:lossderiv_mu}) and we know
\begin{align}
\frac{\partial \Phi_{i}}{\partial x_{i}} &= \frac{\sqrt{2} \exp\left(-\frac{{\left(\mu_{i} - x_{i}\right)}^{2}}{2 \, \sigma_{i}^{2}}\right)}{2 \, \sqrt{\pi} \sigma_{i}}, \label{eqn:pder_x}
\end{align}

\noindent
Then, $\frac{\partial \ell}{\partial x_{i}}$  is obtained
\begin{equation}
\resizebox{1.0\linewidth}{!}{$
\frac{\partial \ell}{\partial x_{i}}  = \left( \left( \left(x_{i} + \lambda \Phi_{i}\right)\sum_{j}^{}y_{j} \right) - y_{i} \right) \left(1 + \lambda\frac{\sqrt{2} \exp\left(-\frac{{\left(\mu_{i} - x_{i}\right)}^{2}}{2 \, \sigma_{i}^{2}}\right)}{2 \, \sqrt{\pi} \sigma_{i}} \right).
$}
\label{eqn:lossder_final_x}
\end{equation}

\subsection{$\mathcal{G}$-Softmax in Multi-label Classification}
\label{subsec:multilabel}

Subsection \ref{sec:sub1} is based on the single-label classification problems. Here, we apply the proposed $\mathcal{G}$-Softmax function to the multi-label classification problem. In the single-label classification problems, the softmax loss and the $\mathcal{G}$-softmax variant are defined as
\begin{small} 
	\begin{align}
	\begin{split}
	\text{Softmax: }\ell &=  -\sum_{i=1}^{m} y_{i} \log\left(\frac{\exp(x_{i})}{\sum_{j=1}\exp(x_{j})}\right), \\
	\mathcal{G}\text{-Softmax: }\ell &=  -\sum_{i=1}^{m} y_{i} \log\left(\frac{\exp\left(x_{i} + \lambda \Phi(x_{i}; \mu_{i}, \sigma_{i}) \right)}{\sum_{j=1} \exp\left(x_{j} + \lambda \Phi(x_{j}; \mu_{j}, \sigma_{j}) \right)} \right).
	\end{split}
	\label{eqn:sgl_loss}
	\end{align}
\end{small}

For multi-label classification, multi-label soft margin loss (MSML) is widely used to solve the multi-label classification problems \cite{Durand_CVPR_2017, Zhu_CVPR_2017}, as defined by Equation (\ref{eqn:msml}). 
\begin{align}
\begin{split}
\ell =  -\sum_{i=1}^{m} y_{i} & \log\left(\frac{1}{1+\exp(-x_{i})}\right) \\ &+(1-y_{i}) \log\left(1-\frac{1}{1+\exp(-x_{i})}\right).
\end{split}
\label{eqn:msml}
\end{align}

In contrast with MSML, there is a variant that takes $x_{i}^{+}$ and $x_{i}^{-}$ as inputs, instead of only taking $x_{i}$ as inputs in MSML. $x_{i}^{+}$ is the positive feature which is used to compute the probability that the input image is classified to the $i$-th category, while $x_{i}^{-}$ is the negative feature that is used to compute the probability that the input image is classified to the non-$i$-th category. The variant is used in the multi-label classification problems \cite{Li_MTA_2018}. It is defined by Equation (\ref{eqn:mhml}).

\begin{align}
\begin{split}
\ell =  -\sum_{i=1}^{m} y_{i} & \log\left(\frac{1}{1+\exp(-x_{i}^{+})}\right) \\ &+(1-y_{i}) \log\left(\frac{1}{1+\exp(-x_{i}^{-})}\right).
\end{split}
\label{eqn:mhml}
\end{align}

The terms {\small $1 / \left(1+\exp(-x_{i})\right)$} and {\small $1 - 1 / \left(1+\exp(-x_{i})\right)$} in MSML (\ref{eqn:msml}) are both determined by $x_{i}$. To make the learning process consistent with the loss function used in single-label classification, we use the variant, \ie Equation (\ref{eqn:mhml}), for multi-label classification in this work and denote it as the softmax loss function for consistency. Correspondingly, the $\mathcal{G}$-softmax loss function is defined as
\begin{align}
\begin{split}
&\ell =  -\sum_{i=1}^{m} y_{i} \log\left(\frac{1}{1+\exp(-x_{i}^{+}-\lambda \Phi(x_{i}^{+}; \mu_{i}^{+}, \sigma_{i}^{+}))}\right) \\ &+(1-y_{i}) \log\left(1-\frac{1}{1+\exp(-x_{i}^{-}-\lambda \Phi(x_{i}^{-}; \mu_{i}^{-}, \sigma_{i}^{-}))}\right).
\end{split}
\label{eqn:gmhml}
\end{align}
In this way, we can model the distributions of $\{x_{i}^{+}\}$ and $\{x_{i}^{-}\}$ by $(\mu_{i}^{+},\sigma_{i}^{+})$ and $(\mu_{i}^{-},\sigma_{i}^{-})$, respectively.

We can see that the proposed $\mathcal{G}$-softmax and softmax function are both straightforward to extend for multi-label classification. In contrast, L-softmax function may not be easy to adapt to multi-label classification. This is because L-softmax function needs to be aware of the feature related to the ground truth label so that it is able to impose a margin constraint on the feature, \ie $p = \frac{\exp(\|W_{y}\|\|x\|\cos(\tilde{m}\theta_{y}))}{\exp(\|W_{y}\|\|x\|\cos(\tilde{m}\theta_{y}))+\sum_{j\ne y}^{} \exp(\|W_{j}\|\|x\|\cos(\theta_{j}))}$, where $\tilde{m}$ is an integer representing the margin, $y$ indicates the $y$-th label is the ground truth label of $x$, $W_{y}$ is the $y$-th column of $W$, and $\theta_{y}$ is the angle between $W_{y}$ and $x$. When $j\ne y$, the exponential term is the same as softmax function. However, when $j = y$, $\tilde{m}$ is used to guarantee the margin between $\|W_{y}\|\|x\|\cos(\tilde{m}\theta_{y})$ and $\|W_{j}\|\|x\|\cos(\theta_{j})$ $(j\ne y)$. As a consequence, it is hard to use in MSML because L-softmax function will treat the terms in Equation (\ref{eqn:msml}) differently.

\subsection{Malleable Learning Rates}
The training of a model usually required a series of predefined learning rates. The learning rate is a real value and a function of the current epoch with given starting and final value. There are several popular types of learning rates, \eg linspace, logspace, and staircase. Usually, the number of epochs with these types of learning rates is not more than 300. Although Huang \etal~\cite{Huang_ICLR_2017} use many more epochs with annealing learning rates, the learning rate is designed as a function of iteration number instead of epoch number. Therefore, it may not generalize to distributed or parallel processing because the iterations are not processed sequentially. We would like to test the proposed $\mathcal{G}$-Softmax function for an extreme condition, \ie more epochs, to investigate the stability. In the following, we first describe the three learning rates followed by showing how these learning rates are in correlation to the proposed malleable learning rate. The proposed malleable learning rates can control the curvature of the scheduled learning rates to boost convergence of the learning process.

The linspace learning rates are generated with a simple linear function, where learning rate at $n$ epoch, {\small $\eta^{(n)}$}, is denoted as {\small $\eta^{(n)} = (a+\frac{b-a}{M-1}(n-1)) \times \eta^{(0)}$}. Here, $M$ is the maximum epoch number, while $a$ and $b$ are the starting and final value of the learning sequences, respectively. {\small $\eta^{(0)}$} is the initial learning rate. Because of linearity, changes of the learning rates are constant through all epochs. As the learning rates become smaller when epoch number increase, it is expected that the training process can converge stably. Logspace learning rates meet this requirement by a log function  {\small $\eta^{(n)} = \exp(\log(a)+\frac{\log(b)-\log(a)}{M-1}(n-1)) \times \eta^{(0)}$}. 

The logspace learning rate has a gradual descent trace that rapidly becomes stable. On the other hand, the staircase learning rate remains constant for a large number of epochs. As the learning rate is not frequently adjusted, the model learning process may not converge. These problems undermine the sustainable convergence ability of deep learning model. Therefore, we integrate the advantages of these learning rates and propose a malleable learning rate, that is,

{\footnotesize
\begin{equation}
	\begin{split}
	\eta^{(n)} \!\! =
	\begin{cases}
	\exp(\log(a_{1})+\frac{\log(b_{1})-\log(a_{1})}{M-1}(n-1)) \! \times \! \eta^{(0)}, \!\!\! & \text{$n\le n_{1}$} \\
	\exp(\log(a_{2})+\frac{\log(b_{2})-\log(a_{2})}{M-1}(n-1)) \! \times \! \eta^{(0)}, \!\!\! & \text{$n_{1} < n\le n_{2}$} \\
	\ldots\\
	\exp(\log(a_{n})+\frac{\log(b_{n})-\log(a_{n})}{M-1}(n-1)) \! \times \! \eta^{(0)}, \!\!\! & \text{$n\le N$} \\
	\end{cases},
	\end{split}
	\label{eqn:learningcurve}
\end{equation}
}%
where $n_{i}$ is the end epoch of the $i$-th piece of learning rates and $a_{n} = b_{n-1}$. As shown in Equation~(\ref{eqn:learningcurve}), the propose learning rate is able to separate piece wise learning rates (\ie staircase learning rates), yet able to control the shape of each piece (\eg curvature or degree of bend) by configuring $a_{i}$ and $b_{i}$.

For the experiments using pre-trained models with the ImageNet dataset \cite{Russakovsky_IJCV_2015}, the initialization contains well learned knowledge for Tiny ImageNet, MS COCO, and NUS-WIDE, which are similar to ImageNet in terms of visual content and concept labels. Hence, the training process on these datasets do not need a number of epochs \cite{Zhu_CVPR_2017, Durand_CVPR_2017}. In this work, we instead apply malleable learning rates on CIFAR to train the models from scratch.

\subsection{Compactness \& Separability}
\label{sec:subseparability}

As commonly studied in machine learning~\cite{Liu_ICML_2016,Yang_AAAI_2006,Zhang_ICML_2007}, intra-class compactness and inter-class separability are important characteristics that can reveal some intuition about the learning ability and efficacy of a model. Due to the underlying Gaussian nature of the proposed $\mathcal{G}$-softmax function, the intra-class compactness for a given class $c$ is characterized by the respective standard deviation $\sigma_c$, where smaller $\sigma_c$ indicates the learned model is more compact. Mathematically, the compactness of a given class $c$ can be represented by $\frac{1}{\sigma_{c}}$.

The inter-class separability can be measured by computing the disparity of two models, \ie the divergence between two Gaussian distributions. 
In the probability and information theory literature, KLD is commonly used to measure the difference between two probability distributions. In the following, we denote a learned Gaussian distribution {\small $\mathcal{N}_i(\mu_{i}, \sigma_{i}^{2})$} as {\small $\mathcal{N}_i$}. Specifically, given two learned Gaussian distributions {\small $\mathcal{N}_{i}$} and {\small $\mathcal{N}_{j}$}, the divergence between two distributions is
\begin{equation}
\resizebox{0.88\linewidth}{!}{$
\begin{split}
\mathcal{D}_{KL} (\mathcal{N}_i\|\mathcal{N}_j) = & -\!\! \int \!\! \phi_{i}(x)\log(\phi_{j}(x))dx + \!\! \int \!\! \phi_{i}(x)\log(\phi_{i})dx \\
 =&\log\frac{\sigma_{j}}{\sigma_{i}} + \frac{\sigma_{i}^{2}+(\mu_{i}-\mu_{j})^{2}}{2\sigma_{j}^{2}}-\frac{1}{2},
\end{split}
\label{eqn:seperability}
$}
\end{equation}

\noindent
where $\phi_{i}$ and $\phi_{j}$ are the probability density functions of the respective class. KLD is always non-negative. As proven by Gibbs' inequality, KLD is zero if and only if the two distributions are equivalent almost everywhere. To quantify the divergence $d_{i}$ between the distribution of the $i$-th category and the distributions of the rest of categories, we use the mean of KLDs,
\begin{equation}
d_{i} = \frac{1}{2(m-1)}\sum_{j\ne i}\left(\mathcal{D}_{KL} (\mathcal{N}_i\|\mathcal{N}_j) + \mathcal{D}_{KL} (\mathcal{N}_j\|\mathcal{N}_i)\right).
\label{eqn:seperability_mean}
\end{equation}

\noindent
Because KLD is asymmetric, we compute the mean of {\small $\mathcal{D}_{KL} (\mathcal{N}_i\|\mathcal{N}_j)$} and {\small $\mathcal{D}_{KL} (\mathcal{N}_j\|\mathcal{N}_i)$} for a fair measurement.

Since compactness indicates the intra-class correlations and separability indicates the inter-class correlations, we multiply (which is the $\times$ operator) intra-class compactness with inter-class separability to overall quantify how discriminative the features with the same label are. Hence we define separability-$\sigma$ ratio $r$, with respect to the $i$-th class as follows
\begin{equation}
r_{i} = \text{separability}\times \text{compactness} = \frac{d_{i}}{\sigma_{i}}.
\label{eqn:s-c-ratio}
\end{equation}

\noindent
Since $\sigma$ of a distribution is inversely proportional to compactness, $r_{i}$ is also inversely proportional to $\sigma$. Ideally, we hope a model's $r$ is as large as possible, which requires separability as large as possible and $\sigma$ as small as possible at the same time.

%% file: depd/experiment.tex
\section{Empirical Evaluation}
\label{sec:experiment}

In this section, we provide comprehensive comparison between the softmax function and the proposed $\mathcal{G}$-softmax function for single-label classification and multi-label classification. Specifically, we evaluate three baseline ConvNets (\ie VGG, DenseNet, and wide ResNet) on CIFAR-10 and CIFAR-100 datasets for single-label classification. For multi-label classification, we conduct the experiments with ResNet on the MS~COCO dataset.

\noindent\textbf{Datasets \& Evaluation Metrics}. To evaluate the proposed $\mathcal{G}$-softmax function for single-label classification, we use the CIFAR-10~\cite{Krizhevsky_Citeseer_2009} and CIFAR-100 datasets, which are widely used in machine learning literature~\cite{Lin_ICLR_2014, Lee_AISTATS_2015, Springenberg_ICLR_2015, Liu_ICML_2016, Clevert_ICLR_2016, Lee_AISTATS_2016, Zagoruyko_BMVC_2016, Huang_CVPR_2017}. CIFAR-10 consists of 60,000 color images with 32$\times$32 pixels in 10 classes. Each class has 6,000 images including 5,000 training images and 1,000 test image. CIFAR-100 has 100 classes and the image resolution is same as CIFAR-10. It has 600 images per class including 500 training images and 100 test images. Moreover, we also use Tiny ImageNet in this work. It is a variant of ImageNet, which has 200 classes and each class has 500 training images and 50 validation images.

For multi-label classification task, we adopt widely used datasets, \ie MS~COCO~\cite{Lin_ECCV_2014} and NUS-WIDE~\cite{Chua_CIVR_2009}.
The MS~COCO dataset is primarily designed for object detection in context, and it is also widely used for multi-label recognition. Therefore, MS~COCO is adopted in this work. It comprises a training set of 82,081 images, and a validation set of 40,137 images. The dataset covers 80 common object categories, with about 3.5 object labels per image. In this work, we follow the original split for training and test, respectively. Following \cite{Li_CVPR_17, Zhu_CVPR_2017, Wang_2017_ICCV, Durand_CVPR_2017}, we only use the image labels for training and evaluation. NUS-WIDE consists of 269,648 images with 81 concept labels. We use official train/test split \ie 161,789 images for training and 107,859 images for evaluation.

We use the same evaluation metrics as \cite{Zhu_CVPR_2017,Wang_2017_ICCV}, namely mean average precision (mAP), per-class precision, recall, F1 score (denoted as C-P, C-R, C-F1), and overall precision, recall, F1 score (denoted as O-P, O-R, O-F1). More concretely, average precision is defined as follows
\begin{align}
AP_{i} = \frac{\sum_{k=1}^{R}\hat{P}_{i}(k)rel_{i}(k)}{\sum_{k=1}^{R}rel_{i}(k)}.
\end{align}
where $rel_{i}(k)$ is a relevant function that returns 1 if the item at the rank $k$ is relevant to the $i$-th class and returns 0 otherwise.
To compute mAP, we collect all predicted probabilities for each class of all the images. The corresponding predicted $i$-th labels over all images are sorted in descending order. The average precision of the $i$-th class is the average of precisions predicted correctly $i$-th labels. $\hat{P}_{i}(k)$ is the precision ranked at $k$ over all predicted $i$-th labels. $R$ denotes the number of predicted $i$-th labels. Finally, the mAP is obtained by averaging AP over all classes. The other metrics are defined as follows
\begin{align}
\begin{split}
\text{C-P} &=\frac{1}{C} \sum_{i}^{}\frac{N_{i}^{c}}{N_{i}^{p}} \hspace{7.1ex} \text{O-P} =\frac{\sum_{i}^{} N_{i}^{c}}{\sum_{i}^{} N_{i}^{p}} \\
\text{C-R} &=\frac{1}{C} \sum_{i}^{}\frac{N_{i}^{c}}{N_{i}^{g}} \hspace{7ex} \text{O-R} =\frac{\sum_{i}^{} N_{i}^{c}}{\sum_{i}^{} N_{i}^{g}} \\
\text{C-F1} &=2 \ \frac{\text{C-P}\times \text{C-R}}{\text{C-P}+\text{C-R}} \hspace{4.9ex} \text{O-F1} =2 \ \frac{\text{O-P}\times \text{O-R}}{\text{O-P}+\text{O-R}} \\
\end{split}
\label{eqn:mc_metrics}
\end{align}

\noindent 
where $N_{i}^{c}$ is the number of images that correctly predicted for the $i$-th class, $N_{i}^{p}$ is the number of predicted images for the $i$-th label, $N_{i}^{g}$ is the number of ground truth images for the $i$-th label. For C-P, C-R, and C-F1, $C$ is the number of labels.

\noindent\textbf{Baselines \& Experiment Configurations}. 
For the classification task, we adopt softmax and L-softmax \cite{Liu_ICML_2016} as baseline methods for comparison purposes. For multi-label classification, due to the limits of L-softmax as discussed in Subsection \ref{subsec:multilabel}, we only use softmax as the baseline method.

There are a number of ConvNets, such as AlexNet~\cite{Krizhevsky_NIPS_2012}, GoogLeNet~\cite{Szegedy_CVPR_2015}, VGG~\cite{Simonyan_ICML_2015}, ResNet~\cite{He_CVPR_2016}, wide ResNet~\cite{Zagoruyko_BMVC_2016}, and DenseNet~\cite{Huang_CVPR_2017}. 
For the experiment on CIFAR-10 and CIFAR-100, we adopt the state-of-the-art wide ResNet and DenseNet as baseline models. Also, considering that the network structure of wide ResNet and DenseNet are quite different with conventional networks, such as AlexNet and VGG, VGG is taken into account too. Specifically, we use VGG-16 (16-layer model), wide ResNet with 40 convolutional layers and the widening factor 14, and DenseNet with 100 convolutional layers and the growth rate 24 in this work.
Our experiments focus on comparing the conventional softmax function with the proposed $\mathcal{G}$-softmax function. Softmax and L-softmax function are considered the baseline functions in this work. For fair comparisons, the experiments are strictly conducted under the same conditions. For all comparisons, we only replace the softmax function in the final layer with the proposed $\mathcal{G}$-softmax function, and preserve other parts of the network. In the training stage, we keep most of training hyperparameters, \eg weight decay, momentum and so on, the same as AlexNet \cite{Krizhevsky_NIPS_2012}. 
Both the baseline and the proposed $\mathcal{G}$-softmax function would be trained from scratch under the same conditions. In wide ResNet experiments, the batch size for CIFAR-10 and CIFAR-100 are both 128, which is the number used in its original work~\cite{Zagoruyko_BMVC_2016}. In DenseNet experiments, since its graphics memory usage is considerably higher than wide ResNet's, we use 50 as batch size, which leads to fully graphics memory usage for three GPUs. The hardware used in this work are Intel Xeon E5-2660 CPU and GeForce GTX 1080 Ti. All models are implemented with Torch~\cite{Collobert_NIPS_2011}.

We follow the original experimental settings of the baseline models for the training and evaluation of the softmax function and the $\mathcal{G}$-softmax function. For example, in DenseNet~\cite{Huang_CVPR_2017}, Huang \etal train their model in 300 epochs with staircase learning rates. From 1st epoch to 149th epoch, the learning rate is set to $0.1$. From 150th epoch to 224th epoch, it is $0.01$, and the learning rates of the remaining epochs are $0.001$. The wide ResNet model is trained in 200 epochs~\cite{Zagoruyko_BMVC_2016}. The learning rate is initialized to $0.1$, and at 60th, 120th, 160th, it will decrease to $0.02$, $0.004$, $0.0008$, respectively. To make it comparable to DenseNet, we extend the epochs from 200 to 300 and decrease the learning rate at 220th and 260th epoch by multiplying $0.2$. To avoid ad hoc training of hyperparameter settings, we set the weight decay $\epsilon$ and momentum $\gamma$ to be the same as the default hyperparameters in the baselines~\cite{Zagoruyko_BMVC_2016, Huang_CVPR_2017} (\ie {\small $\epsilon=\num{5e-4}, \gamma=0.9$}) for the softmax function and the proposed $\mathcal{G}$-softmax function.

For the experiments on Tiny ImageNet, we adopt wide ResNet \cite{Zagoruyko_BMVC_2016} with 40 convolutional layers and width 14 as the baseline model. Initial learning rate is 0.001 and weight decay is 1e-4. The training process consists of 30 epochs with batch size 80 and the learning will be decreased to its one tenth every 10 epoch. Following \cite{Huang_ICLR_2017,Yamada_ICLR_2018}, we use the ImageNet pre-trained weights as an initialization and the input image will be resized to 224$\times$224 to feed the wide ResNet.

\begin{table}[!t]
	\centering
	\captionof{table}
	{
		\small
		Top 1 error rate ($\%$) of the proposed $\mathcal{G}$-softmax function and other baselines on CIFAR-10 and CIFAR-100. * indicates that malleable learning rates with 1100 epochs are used in the training process (refer to experimental configurations in Section \ref{sec:experiment} for more details).
	}
	\begin{tabular}{lccc}
		\toprule
		& \# Epoch & CIFAR10 & CIFAR100 \\
		\cmidrule(lr){2-2} \cmidrule(lr){3-3} \cmidrule(lr){4-4}
		DSN \cite{Huang_CVPR_2017} 	& ~300 & 3.46 & 17.18 \\
		WRN \cite{Zagoruyko_BMVC_2016} 	& ~200 & 3.80 & 18.30 \\
		L-softmax \cite{Liu_ICML_2016} 	& 80 & 5.92 & 29.53 \\
		\midrule
		VGG 																& ~300 & 5.69 & 25.07 \\
		VGG L-softmax 											& ~300 & 7.79 & 32.89 \\
		VGG $\mathcal{G}$-softmax 													& ~300 & 5.54 & 24.92 \\
		\midrule
		DSN 														& ~300 & 3.77 & 19.25 \\
		DSN L-softmax 											& ~300 & 4.84 & 23.22 \\
		DSN $\mathcal{G}$-softmax 											& ~300 & 3.67 & 18.89 \\
		\midrule
		WRN 														& ~300 & 3.49 & 17.66 \\
		WRN L-softmax 											& ~300 & 4.27 & 20.53 \\
		WRN $\mathcal{G}$-softmax 												& ~300 & 3.36 & 17.41 \\ \midrule
		*WRN 														& 1100 & 3.18 & 17.60 \\
		*WRN L-softmax 														& 1100 & 4.11  & 20.45 \\
		*WRN $\mathcal{G}$-softmax 						& 1100 & \textbf{3.14} & \textbf{17.04} \\
		\bottomrule			
		\label{tbl:perf_train}
	\end{tabular}
	\label{tbl:perf}
	\vspace{-1ex}
\end{table}

\begin{table}[h!]
	\centering
	\caption{Top 1 error rate ($\%$) on the validation set of Tiny ImageNet. Note that we use input image size 224$\times$224 for ResNet-101 in the experiments.}
	\begin{tabular}{lc}
		\toprule
		& Top 1 error ($\%$) \\
		\cmidrule(lr){2-2} 
		Wide-ResNet \cite{Zagoruyko_BMVC_2016}  & 39.63 \\
		Wide-ResNet SE \cite{Huang_ICLR_2017} & 32.90 \\
		DenseNet \cite{Huang_CVPR_2017} &  39.09 \\
		IGC-V2 \cite{Xie_CVPR_2018} & 39.02 \\
		PyramidNet Shackdrop \cite{Yamada_ICLR_2018} & 31.15 \\ \midrule
		ResNet-101 (Input size: 64$\times$64) & 31.66  \\
		ResNet-101 (Input size: 224$\times$224) & 18.36  \\
		ResNet-101 L-Softmax & 17.57  \\
		ResNet-101 $\mathcal{G}$-Softmax ($\mu=-0.05,\sigma=1$) & \textbf{16.86}  \\
		ResNet-101 $\mathcal{G}$-Softmax ($\mu=0.05,\sigma=1$) & 16.95  \\
		ResNet-101 $\mathcal{G}$-Softmax ($\mu=0,\sigma=1$) & 17.04  \\
		ResNet-101 $\mathcal{G}$-Softmax ($\mu=0,\sigma=2$) & 17.29  \\
		ResNet-101 $\mathcal{G}$-Softmax ($\mu=0,\sigma=3$) & 16.96  \\
		\bottomrule
		\label{tbl:timg_perf}
	\end{tabular}
\end{table}

In the experiments with malleable learning rates, 1100 epochs are used in training. There are only two phases throughout the whole training, \ie $1\le n \le 1000$ and $1000< n \le 1100$, where $(a_{1},b_{1}) = (0,-8)$ and $(a_{2},b_{2}) = (-8,-9)$.

Different from the softmax function, the proposed $\mathcal{G}$-softmax function has two learnable parameters (\ie $\mu$ and $\sigma$) and one hyperparameter (\ie $\lambda$). Without loss of generality, $\mu$ and $\sigma$ are initialized with standard Gaussian distribution (\ie to $0$ and $1$). These two parameters would be learned through training by Equation (\ref{eqn:lossder_final_mu}) and (\ref{eqn:lossder_final_sigma}). To determine $\lambda$, we follow the similar rule where we start from $1$ and try the value between $[0,1]$. As mentioned in Section~\ref{sec:method}, the $\mathcal{G}$-softmax function would be equivalent to the softmax function if $\lambda = 0$. In our experiments, $\lambda$ is initialized to 1 for CIFAR-10 and CIFAR-100 experiments with DenseNet. In wide ResNet, $\lambda$ is initialized as $1$ for CIFAR-100 experiments and $0.5$ for CIFAR-10 experiments.

\begin{table*}[!t]
	\centering
	\caption{Performances on the validation set of MS COCO. C-P, C-R, and C-F1 stand for per-class precision, recall, and F-1 measure, respectively. O-P, O-R, and O-F1 stand for overall precision, recall, and F-1 measure, respectively. All the numbers are presented in percentage ($\%$). GAP indicates a global average pooling is used in the last pooling layer while GMP indicates a global max pooling is used in the last pooling layer. ResNet-101 $\mathcal{G}$-softmax uses the same hyperparameters and architecture as ResNet-101 with the softmax function. For simplicity, we denote ResNet-101 with the softmax function as ResNet-101, \textit{ImgSize} as image size, \textit{bz} as batch size, and \textit{lr} as learning rate. Notablely, mAP, C-F1, and O-F1 are more important to measure the performance \cite{Zhu_CVPR_2017,Wang_2017_ICCV}.}
	\vspace{-1ex}
	\begin{tabular}{lccccccc}
		\toprule
		\hspace{70ex} & C-P & C-R & C-F1 & O-P & O-R & O-F1 & mAP \\
		\cmidrule(lr){2-4} \cmidrule(lr){4-7} \cmidrule(lr){8-8}
		
		VGG MCE \cite{Li_MTA_2018}  & - & - & - & - & - & - & 70.2  \\
		Weak sup (GMP) \cite{Oquab_CVPR_2015}  & - & - & - & - & - & - & 62.8  \\
		CNN-RNN \cite{Wang_CVPR_2016}  & 66.0 & 55.6 & 60.4 & 69.2 & 66.4 & 67.8 & -  \\
		RGNN \cite{Zhao_BMVC_2016}  & - & - & - & - & - & - & 73.0  \\
		WELDON \cite{Durand_CVPR_2016}  & - & - & - & - & - & - & 68.8  \\
		Multi-CNN \cite{Zhang_arXiv_2016}  & 54.8 & 51.4 & 53.1 & 56.7 & 58.6 & 57.6 & 60.4  \\
		CNN+LSTM \cite{Zhang_arXiv_2016}  & 62.1 & 51.2 & 56.1 & 68.1 & 56.6 & 61.8 & 61.8  \\
		MCG-CNN+LSTM \cite{Zhang_arXiv_2016}  & 64.2 & 53.1 & 58.1 & 61.3 & 59.3 & 61.3 & 64.4  \\
		RLSD \cite{Zhang_arXiv_2016} & 67.6 & 57.2 & 62.0 & 70.1 & 63.4 & 66.5 & 68.2  \\
		Pairwise ranking \cite{Li_CVPR_17} & 73.5 & 56.4 & - & 76.3 & 61.8 & - & -  \\
		MIML-FCN \cite{Yang_CVPR_17} & - & - & - & - & - & - & 66.2  \\
		RDAR \cite{Wang_2017_ICCV} & 79.1 & 58.7 & 67.4 & 84.0 & 63.0 & 72.0 & 72.2  \\ 
		ResNet-101 (GAP, ImgSize: $224\times 224$, bz=96, lr=1e-3) \cite{Zhu_CVPR_2017} & 80.8 & 63.4 & 69.5 & 82.2 & 68.0 & 74.4 & 75.2  \\
		SRN \cite{Zhu_CVPR_2017} & 81.6 & 65.4 & 71.2 & 82.7 & 69.9 & 75.8 & 77.1  \\
		
		ResNet-101 (GAP, ImgSize: $448\times 448$, unknown bz and lr) \cite{Durand_CVPR_2017} & - & - & - & - & - & - & 72.5  \\
		WILDCAT \cite{Durand_CVPR_2017} & - & - & - & - & - & - & 80.7  \\
		\midrule
		ResNet-101 (GMP, ImgSize: $448\times 448$, bz=16, lr=1e-5) & 81.3 & 70.2 & 74.1 & 81.9 & 74.3 & 77.9 & 80.6   \\ 
		ResNet-101 $\mathcal{G}$-softmax w/ ($\mu=0$, $\sigma=1$) & 82.4 & 69.3 & 74.1 & 83.2 & 73.3 & 77.9 & 80.8   \\ 
		ResNet-101 $\mathcal{G}$-softmax w/ ($\mu=0$, $\sigma=0.5$) & 82.7 & 69.3 & 74.0 & 83.5 & 72.9 & 77.8 & 81.1   \\ 
		ResNet-101 $\mathcal{G}$-softmax w/ ($\mu=0$, $\sigma=5$) & 80.6 & 71.0 & 74.4 & 81.3 & \textbf{74.7} & 77.8 & 80.9   \\ 
		ResNet-101 $\mathcal{G}$-softmax w/ ($\mu=-0.1$, $\sigma=1$) & 81.3 & \textbf{74.5} & \textbf{74.5} & 83.4 & 73.8 & \textbf{78.3} & \textbf{81.3}   \\  
		ResNet-101 $\mathcal{G}$-softmax w/ ($\mu=0.1$, $\sigma=1$) & \textbf{83.2} & 68.6 & 73.7 & \textbf{84.3} & 72.3 & 77.9 & 81.1   \\ 
		\bottomrule 
		\label{tbl:mscoco_perf}
	\end{tabular}
	\vspace{-2ex}
\end{table*}

\begin{table*}[h!]
	\centering
	\caption{Performance on the validation set of NUS-WIDE. The experimental setting is the same as the experiments on MS COCO. All the numbers are presented in percentage ($\%$). Notablely, mAP, C-F1, and O-F1 are more important to measure the performance \cite{Zhu_CVPR_2017,Wang_2017_ICCV}. Specifically, F1 score is more comprehensive than precision and recall because it takes both precision and recall into account for evaluation. Although LSEP \cite{Li_CVPR_17} achieves better C-P and O-P scores, the C-R, O-R, and C-F1 scores of LSEP are the lowest in the table. The proposed $\mathcal{G}$-softmax function achieves better performance in five metrics including mAP, C-F1, and O-F1.
	}
	\begin{tabular}{lccccccc}
		\toprule
		& C-P & C-R & C-F1 & O-P & O-R & O-F1 & mAP \\
		\cmidrule(lr){2-7} \cmidrule(lr){8-8}
		LSEP \cite{Li_CVPR_17} & \textbf{66.7} & 45.9 & 54.4 & \textbf{76.8} & 65.7 & 70.8 & -   \\
		Order-free RNN \cite{Chen_arXiv_2017} & 59.4 & 50.7 & 54.7 & 69.0 & 71.4 & 70.2 & -   \\
		ResNet-101 \cite{Zhu_CVPR_2017} & 65.8 & 51.9 & 55.7 & 75.9 & 69.5 & 72.5 & 59.8   \\
		\midrule
		ResNet-101 & 62.0 & 56.3 & 56.9 & 74.7 & 71.4 & 73.0 & 59.9   \\
		ResNet-101 $\mathcal{G}$-softmax ($\mu=0$, $\sigma=1$) & 62.5 & \textbf{56.5} & \textbf{57.8} & 74.7 & \textbf{71.7} & \textbf{73.2} & 60.3   \\
		ResNet-101 $\mathcal{G}$-softmax ($\mu=0$, $\sigma=2$) & 62.3 & 56.0 & 57.2 & 74.9 & 71.3 & 73.0 & 60.3   \\
		ResNet-101 $\mathcal{G}$-softmax ($\mu=0$, $\sigma=3$) & 62.9 & 55.9 & 57.1 & 74.9 & 71.2 & 73.0 & 60.1   \\
		ResNet-101 $\mathcal{G}$-softmax ($\mu=-0.05$, $\sigma=1$) & 63.2 & 55.8 & 57.1 & 74.9 & 71.3 & 73.1 & 60.0   \\
		ResNet-101 $\mathcal{G}$-softmax ($\mu=0.05$, $\sigma=1$) & 62.0 & 55.9 & 57.3 & 74.9 & 71.4 & 73.1 & \textbf{60.4}   \\
		\bottomrule
		\label{tbl:nus_perf}
	\end{tabular}
\end{table*}

For the experiments on MS~COCO, we refer to the state-of-the-art works \cite{He_CVPR_2016,Durand_CVPR_2017} to set the weight decay and momentum to $1e^{-4}$ and $0.9$, respectively. The model would be trained with the learning rate $1e^{-5}$ in 8 epochs on MS~COCO validation set. In the experiments, we experiment with various initializations of $\mu$ and $\sigma$ to observe how these factors influence the learning. $\lambda$ is initialized as $1$. Since we follow the convention of multi-label classification \cite{Zhu_CVPR_2017,Durand_CVPR_2017}, we use the pre-trained weights to initialize the ConvNet and this is different from the initializations in the experiments on CIFAR-10 and CIFAR-100. This difference enables the model to determine $\mu$ and $\sigma$ in a data-driven way, that is, empirically computing the $\mu$ and $\sigma$ from data with the pre-trained weights. The image size used in this work is the same as the one used in \cite{Durand_CVPR_2017}, \ie $448\times 448$, while the mini-batch size is 16, which is limited by the number of the GPUs.

For the experiments on NUS-WIDE, we use the same experimental setting as the one on MS COCO.

\begin{figure*}[!t]
	\centering
	\resizebox{1.0\linewidth}{!}{
		\begin{tabular}{cccc}	
			\includegraphics[width=0.235\textwidth]{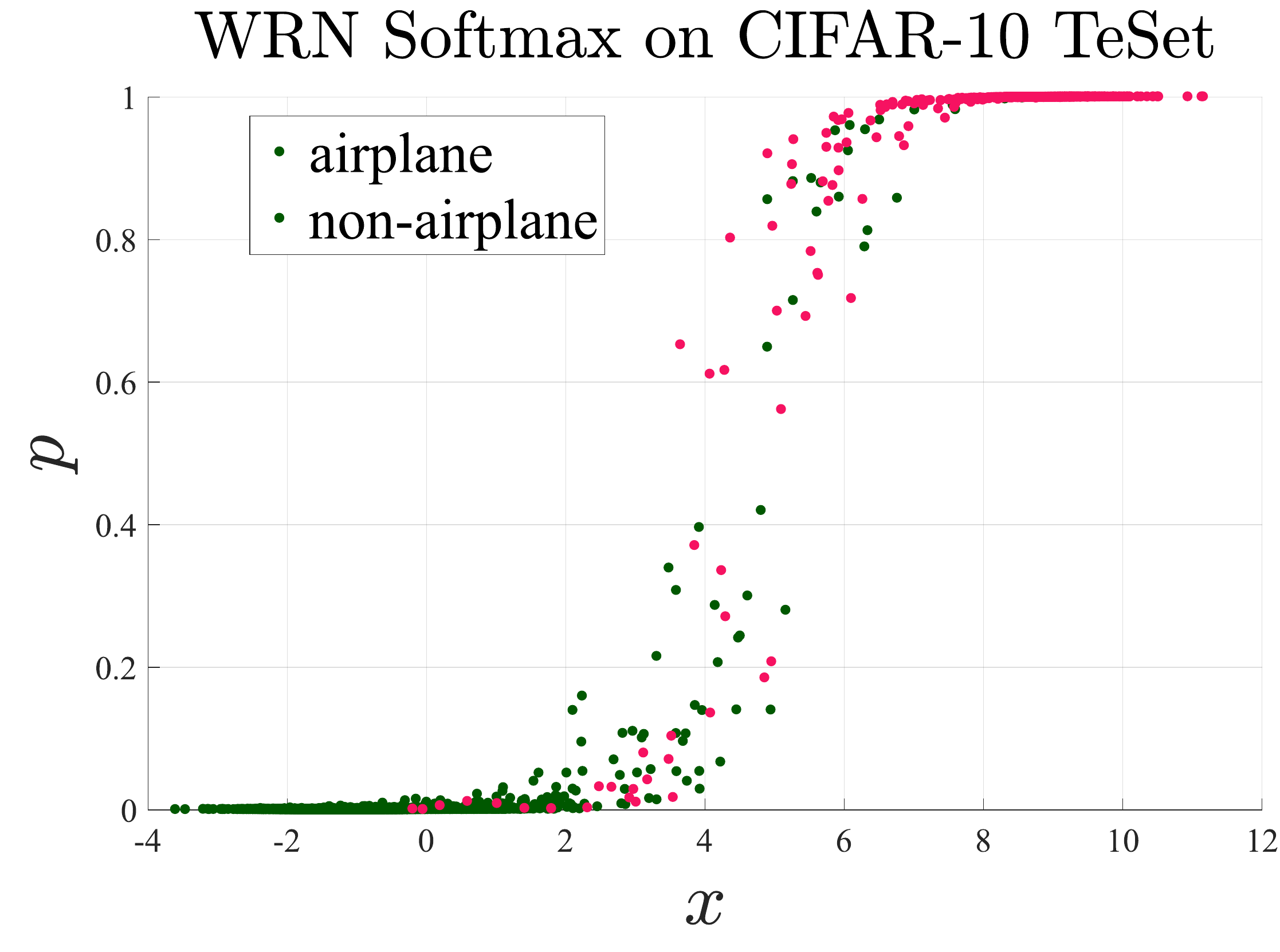} &
			\includegraphics[width=0.235\textwidth]{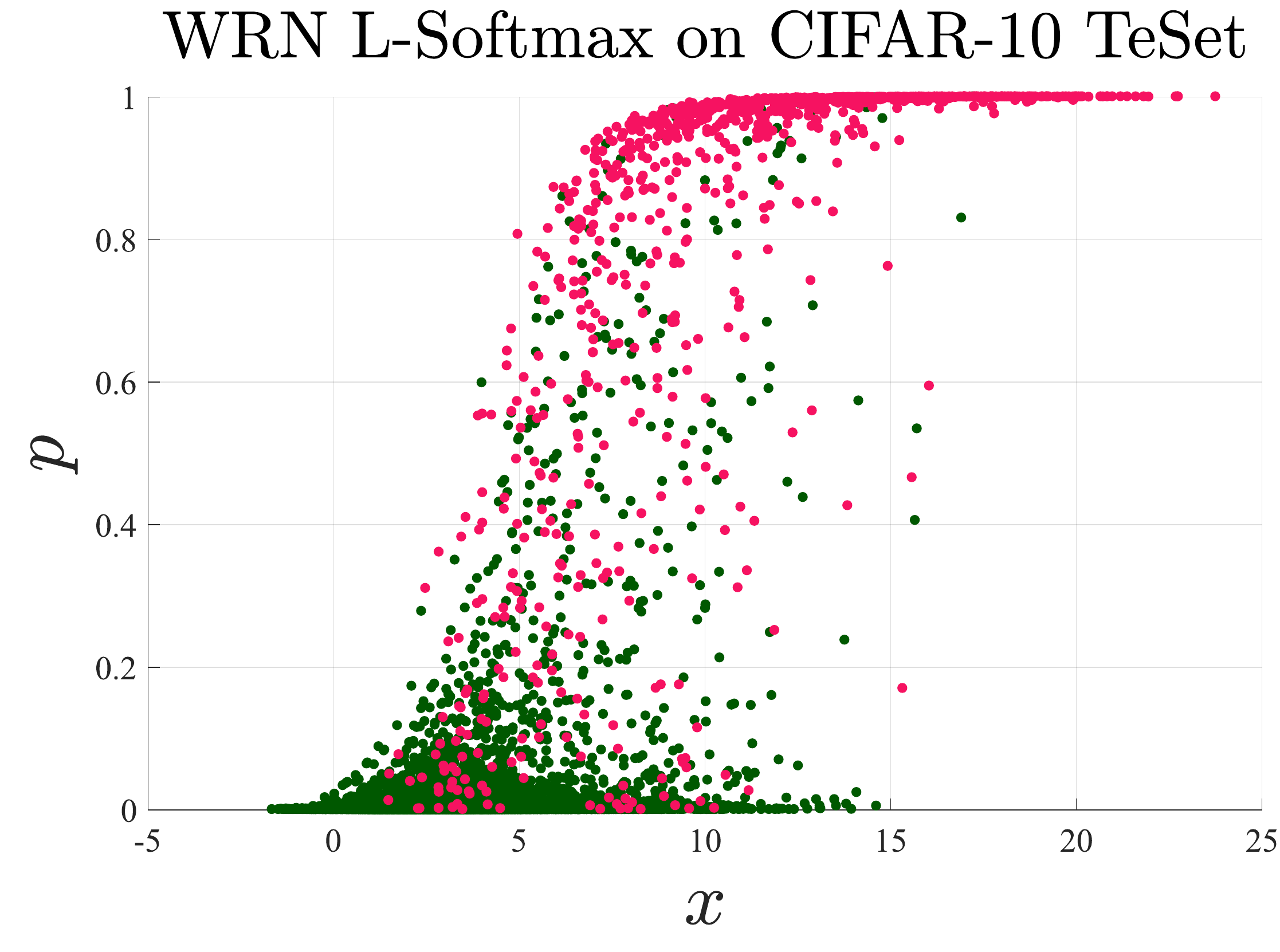} &
			\includegraphics[width=0.235\textwidth]{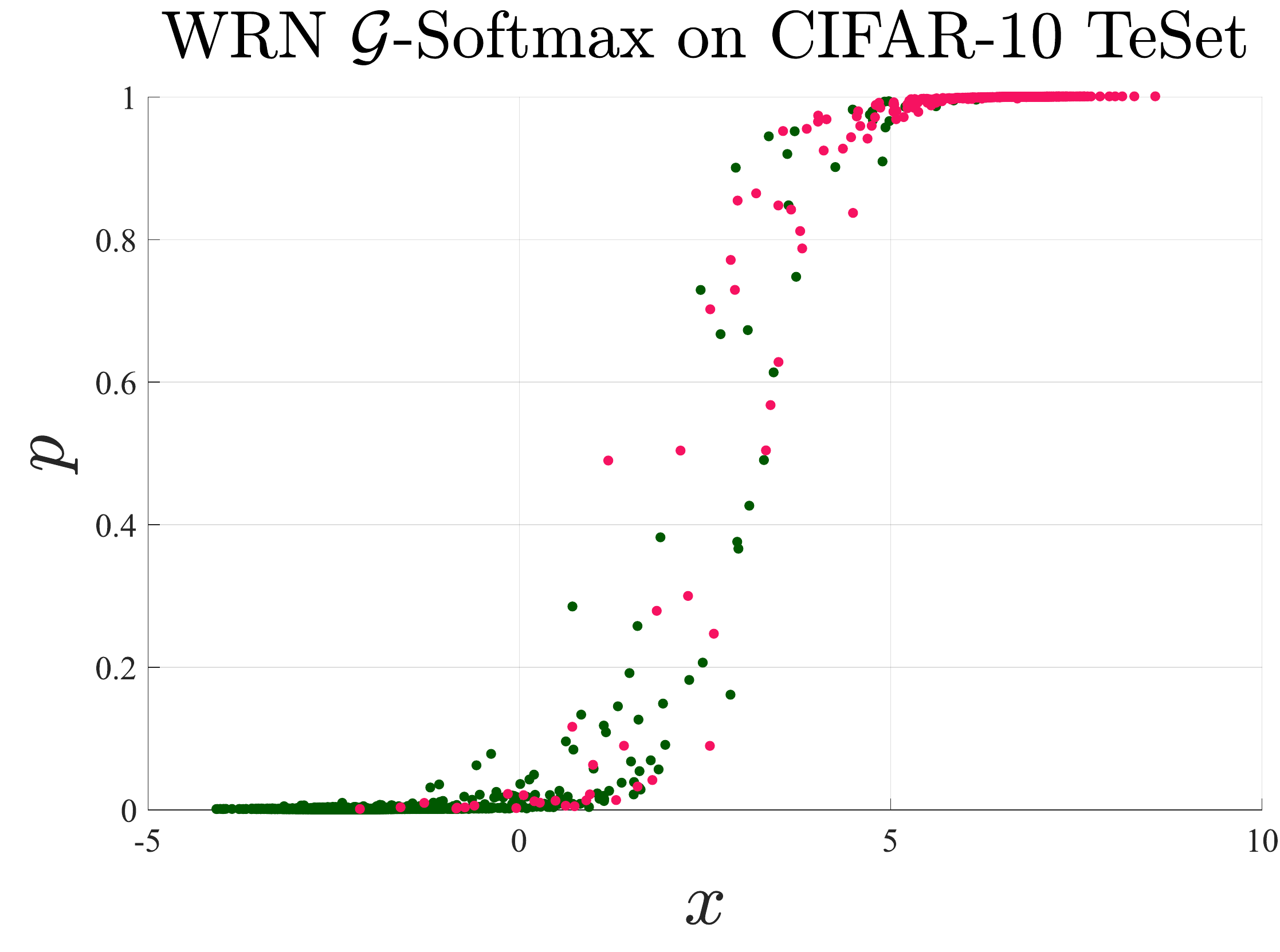} \\
			\includegraphics[width=0.235\textwidth]{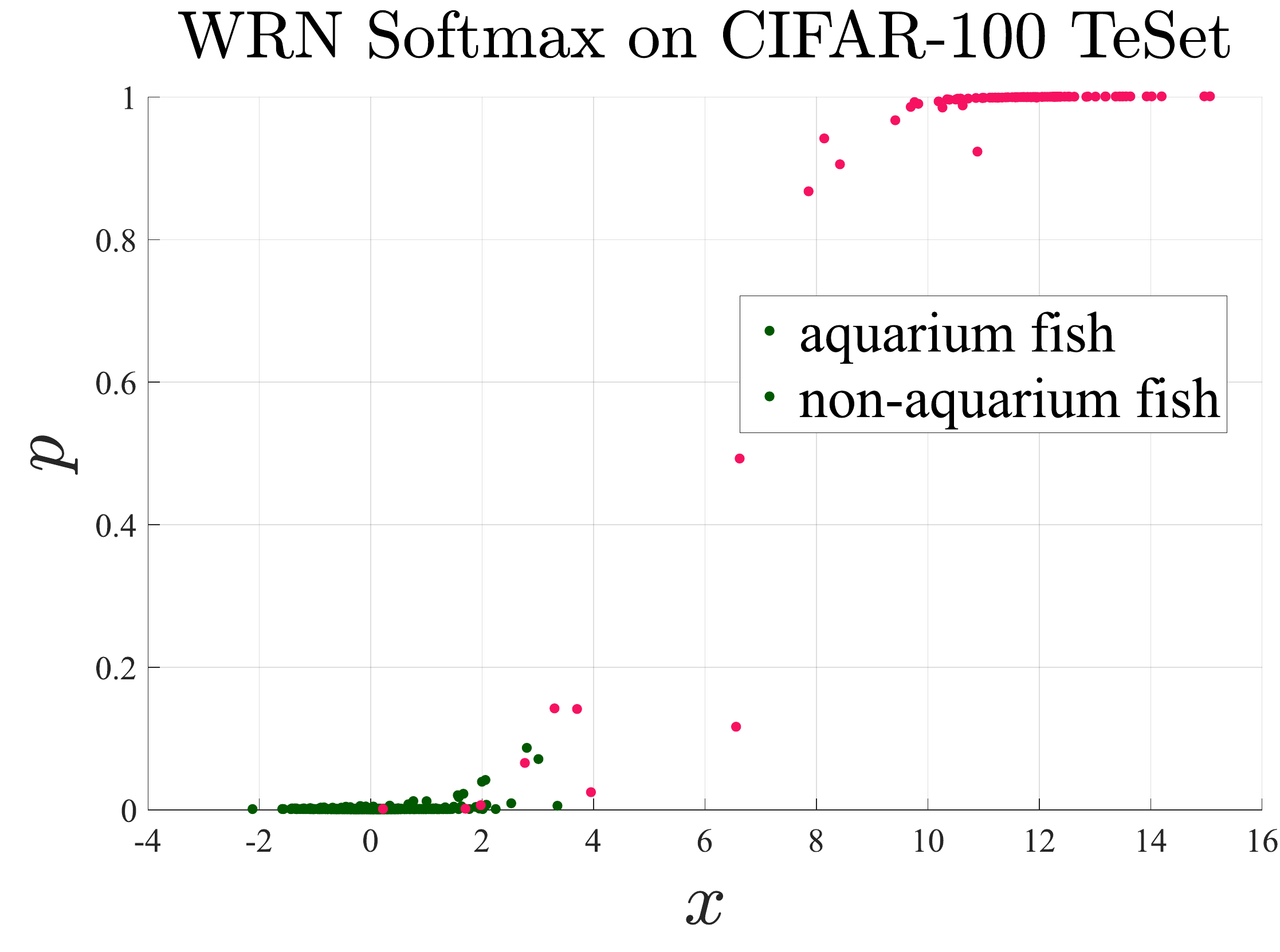} &
			\includegraphics[width=0.235\textwidth]{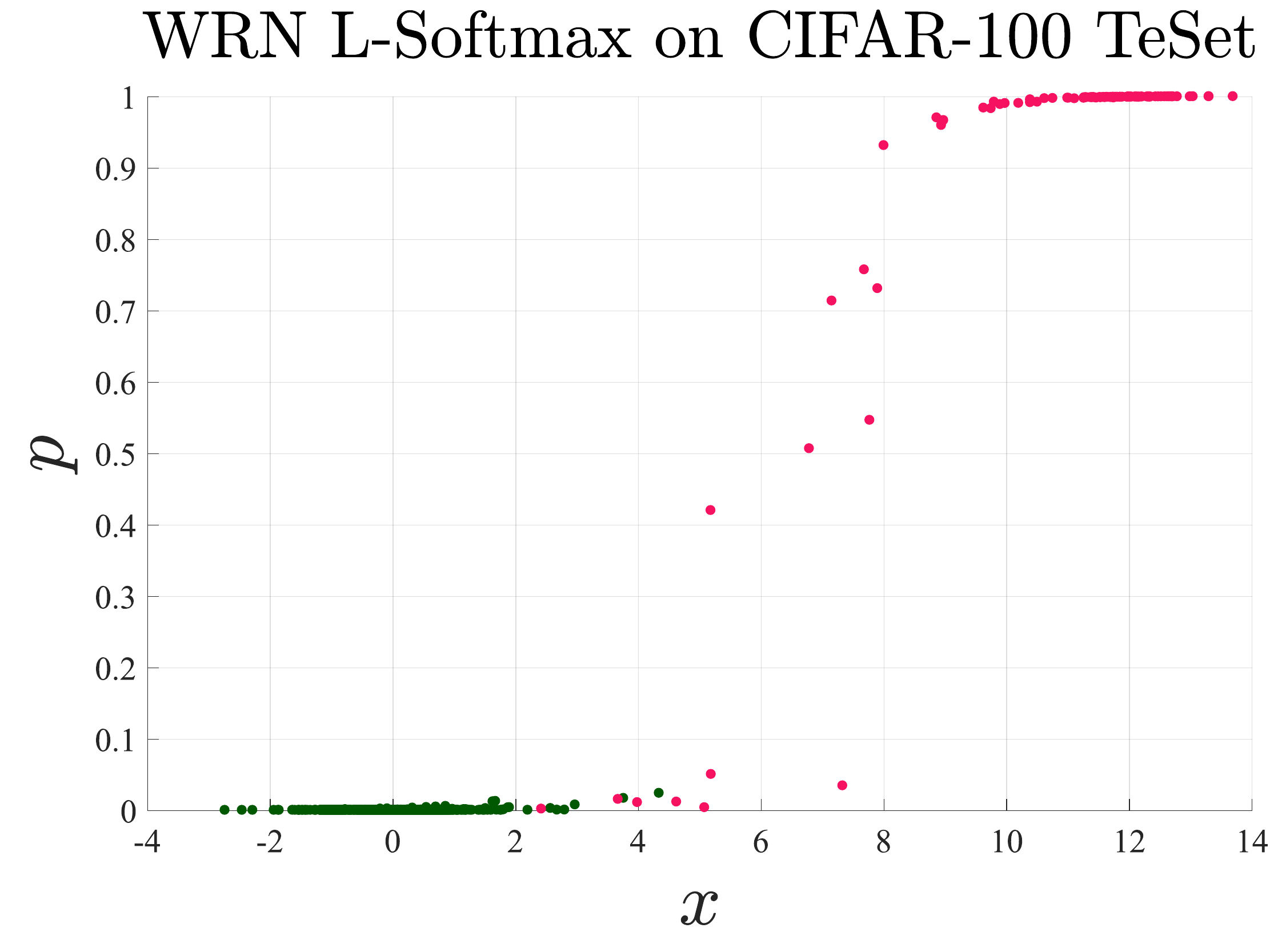} &
			\includegraphics[width=0.235\textwidth]{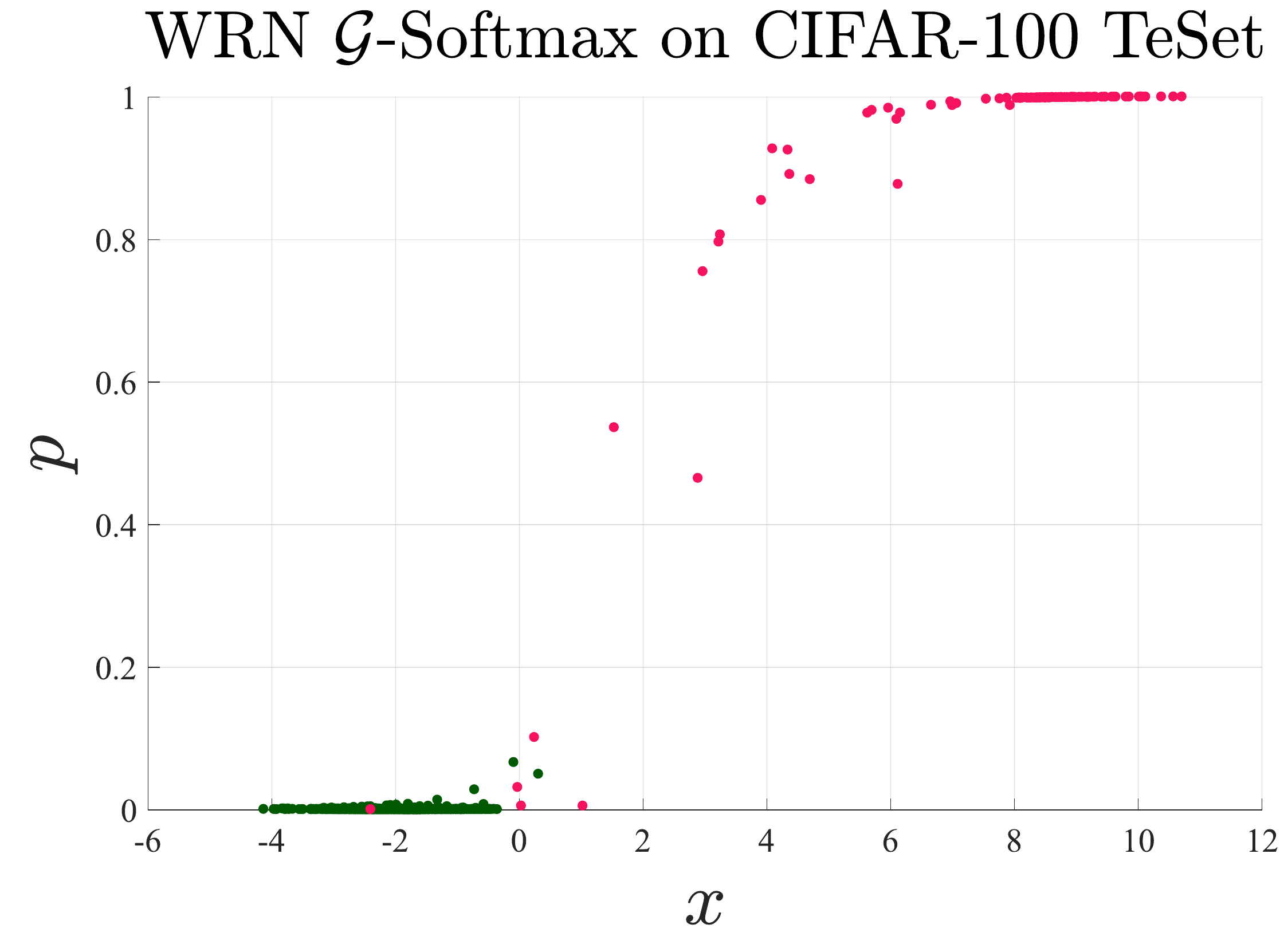} \\
		\end{tabular}
	}
	\vspace{-2ex}
	\caption
	{
		\small
		Prediction vs. feature based on all the test images with the groundtruth class ``airplane'' in CIFAR-10 and ``aquarium fish'' in CIFAR-100, respectively. Softmax, L-softmax, and the proposed $\mathcal{G}$-softmax function are used with wide ResNet for comparison purposes. The 1st row consists of the plots of the experiments on CIFAR-10 while the 2nd row consists of the plots of the experiments on CIFAR-100.  Given all images with the ground truth class ``airplane'', the corresponding ConvNet would extract the deep features $x \in \mathbb{R}^{m}$, $m=10$ in CIFAR-10, and pass them to the predictor for computing the predictions $p$. Here the prediction confidence $p_1$ is corresponding to the ground truth class, where $p_{i\ne1}$ are the predictions as other classes.
		For clarity, we consider all points $(x_{i},p_{i}), i\ne1$ as ``non-airplane'' points and plot them in a scatter plot.
		In this way, the differences of the mapping between the ground truth class and other impostor classes are visualized. In the CIFAR-100 experiments, the same procedures are underwent, but the ground truth class is ``aquarium fish''. Due to space constraint, we only show the results of the first 10 classes in CIFAR-100. 
	}
	\label{fig:scatter}
	\vspace{-2ex}
\end{figure*}

\begin{figure*}[!t]
	\centering
	\begin{tabular}{cccc}
		\includegraphics[width=0.228\textwidth]{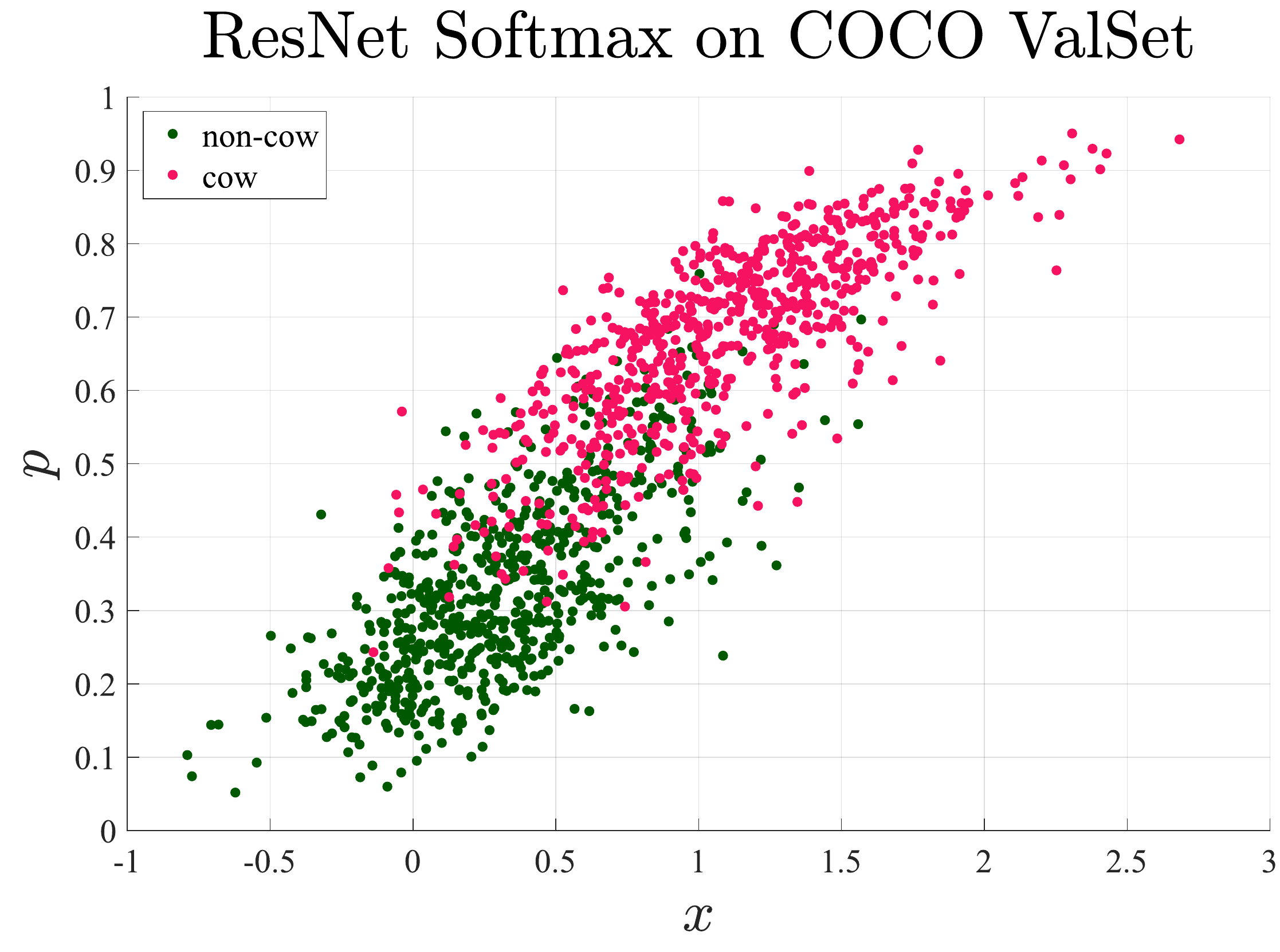} 			 &  \includegraphics[width=0.228\textwidth]{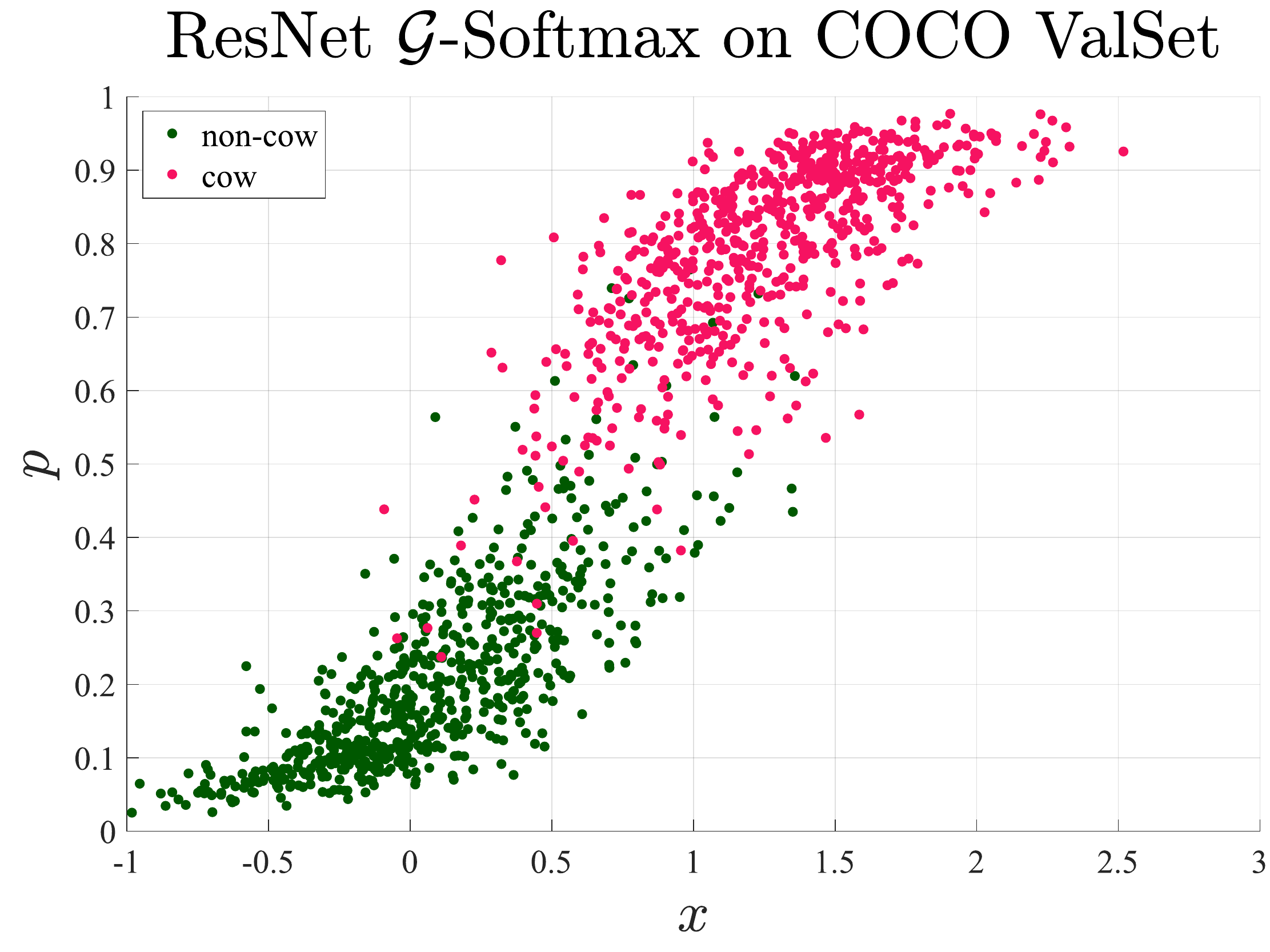} & \includegraphics[width=0.228\textwidth]{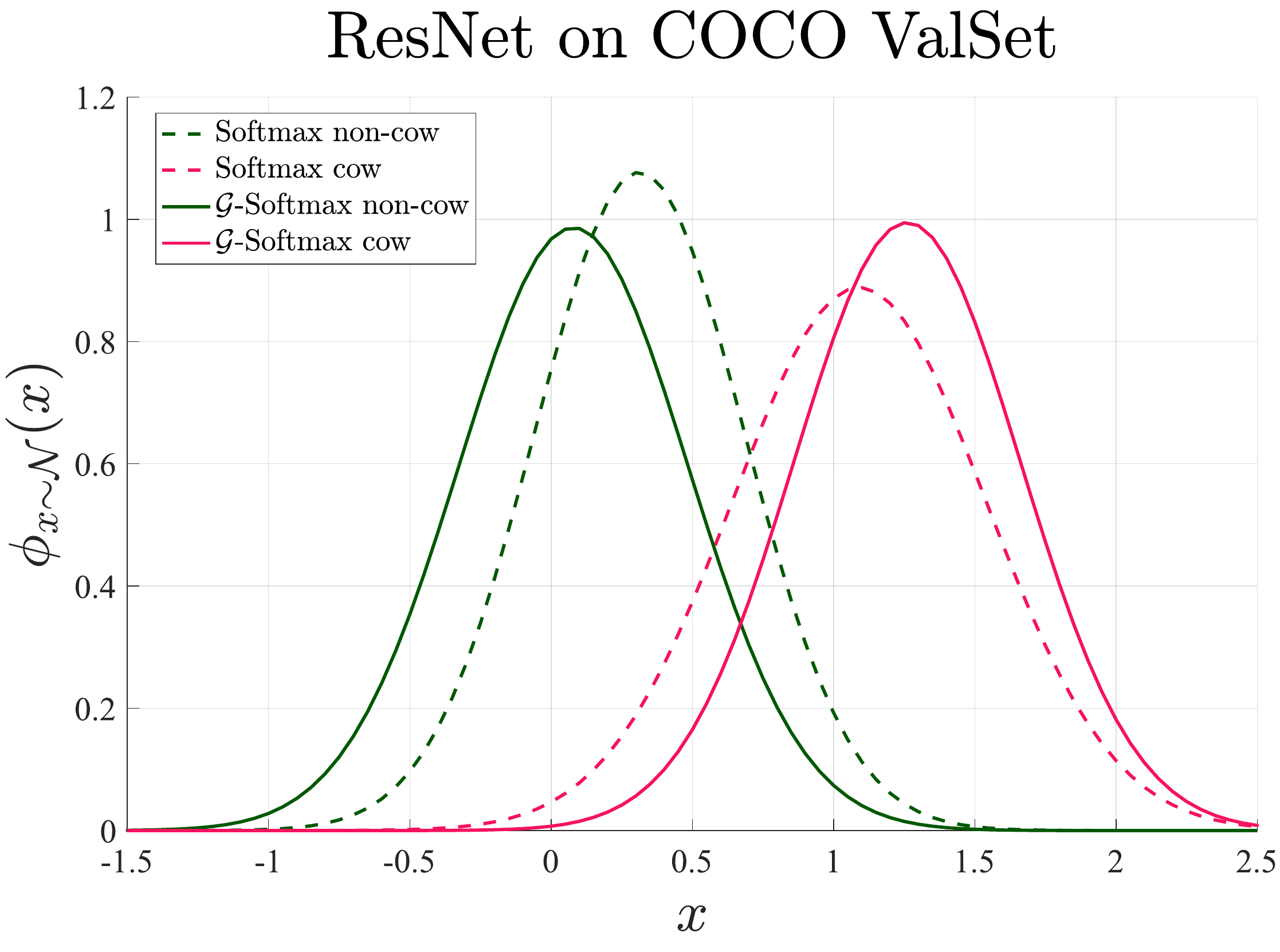} &
		\includegraphics[width=0.228\textwidth]{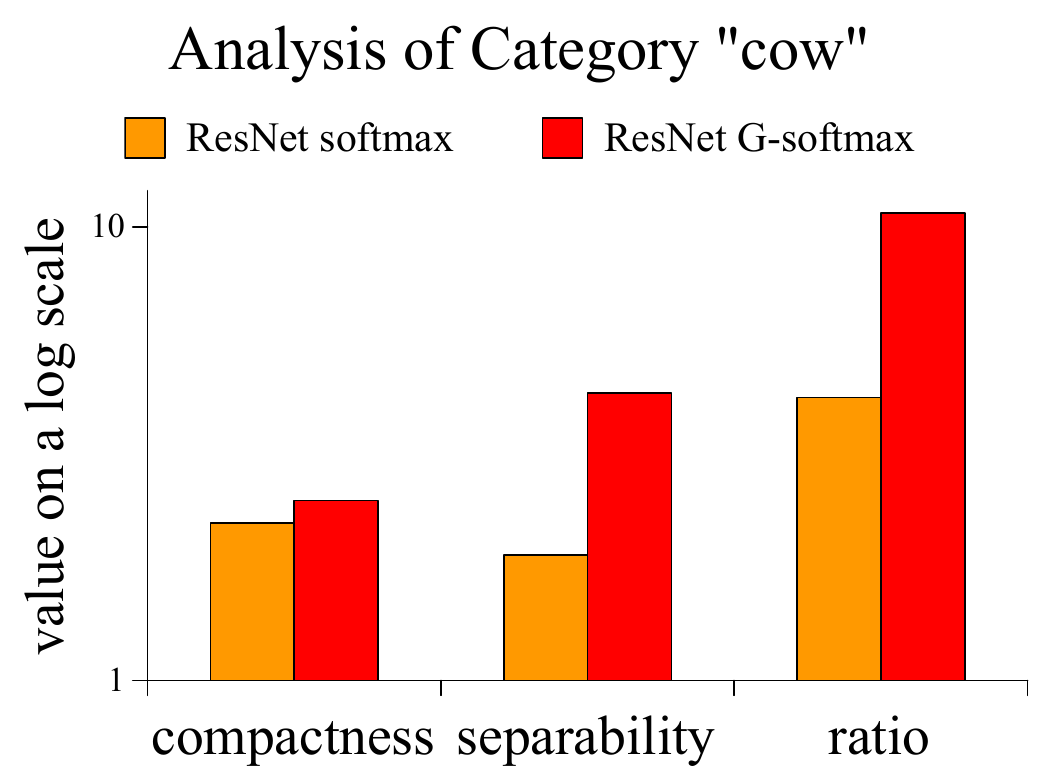} \\
		\includegraphics[width=0.228\textwidth]{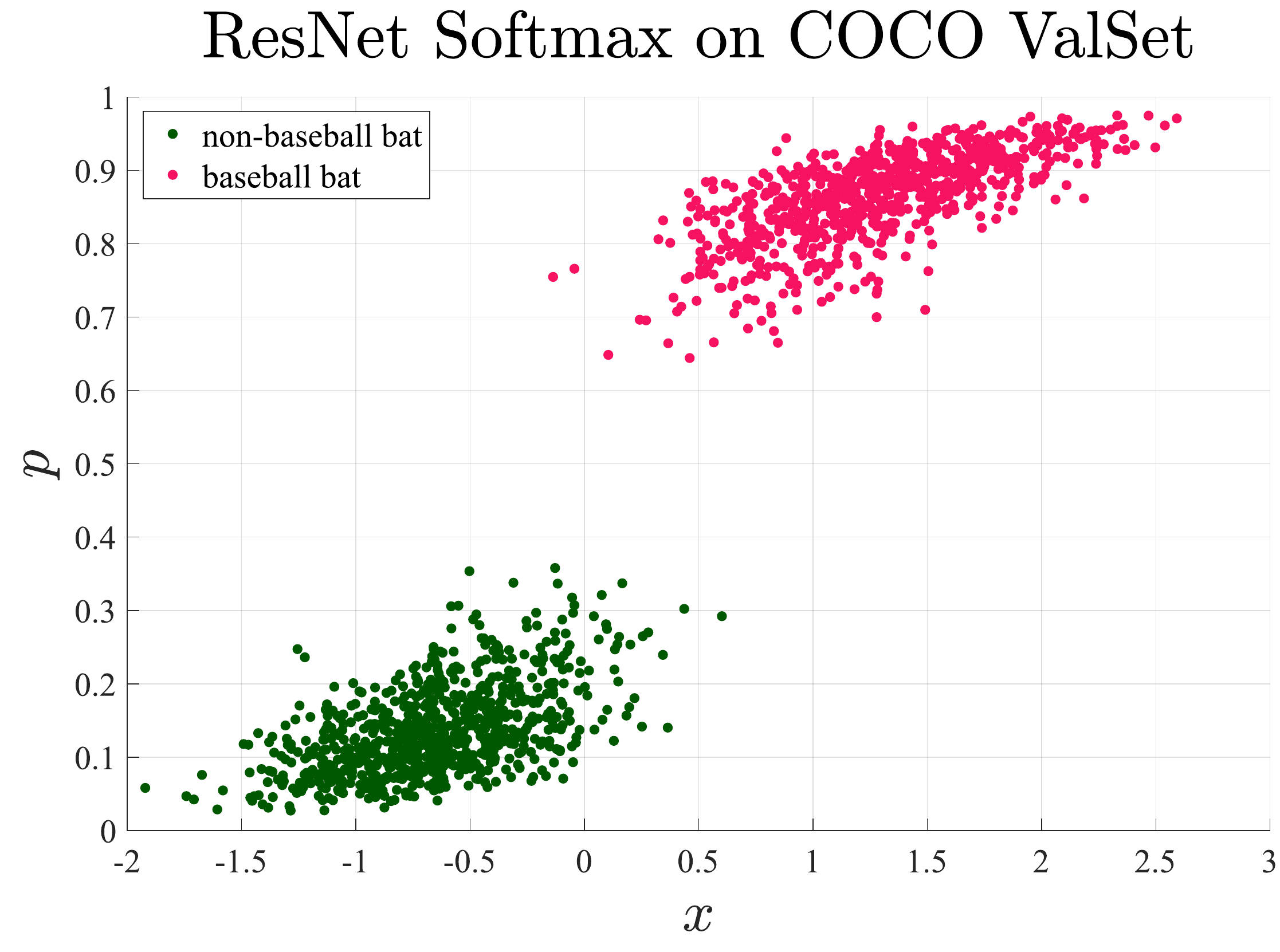} 			 &  \includegraphics[width=0.228\textwidth]{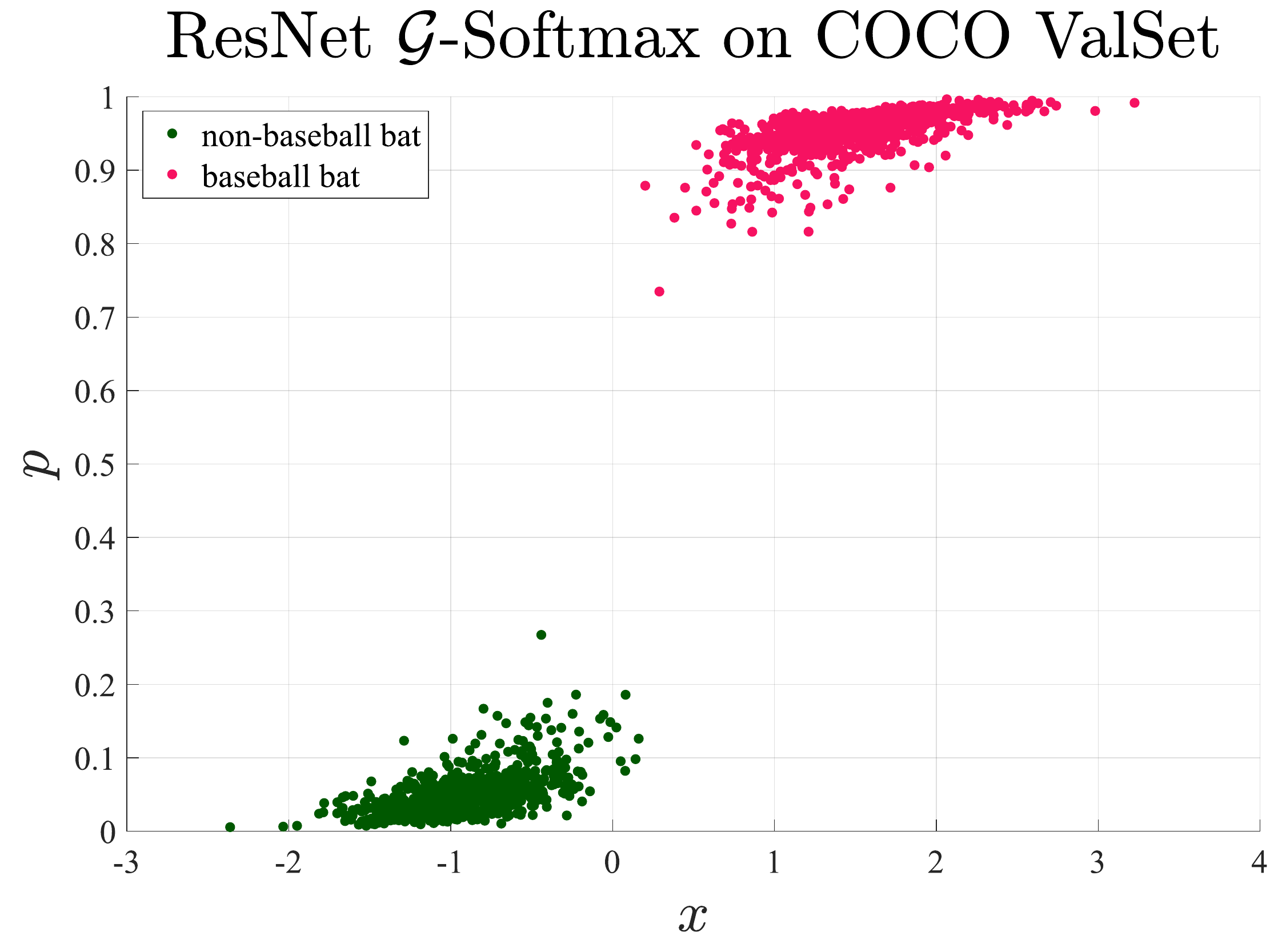} & \includegraphics[width=0.228\textwidth]{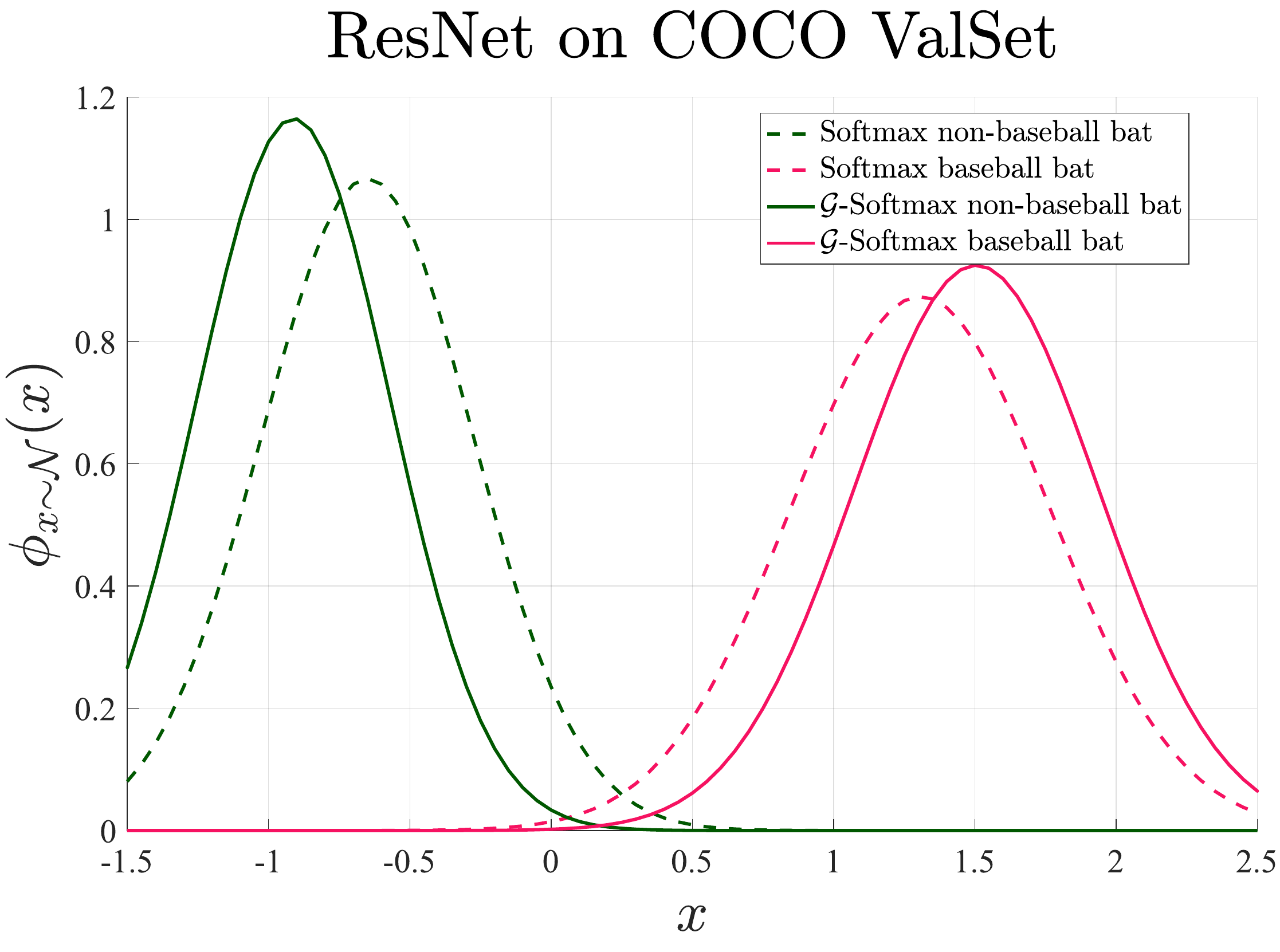}  &
		\includegraphics[width=0.228\textwidth]{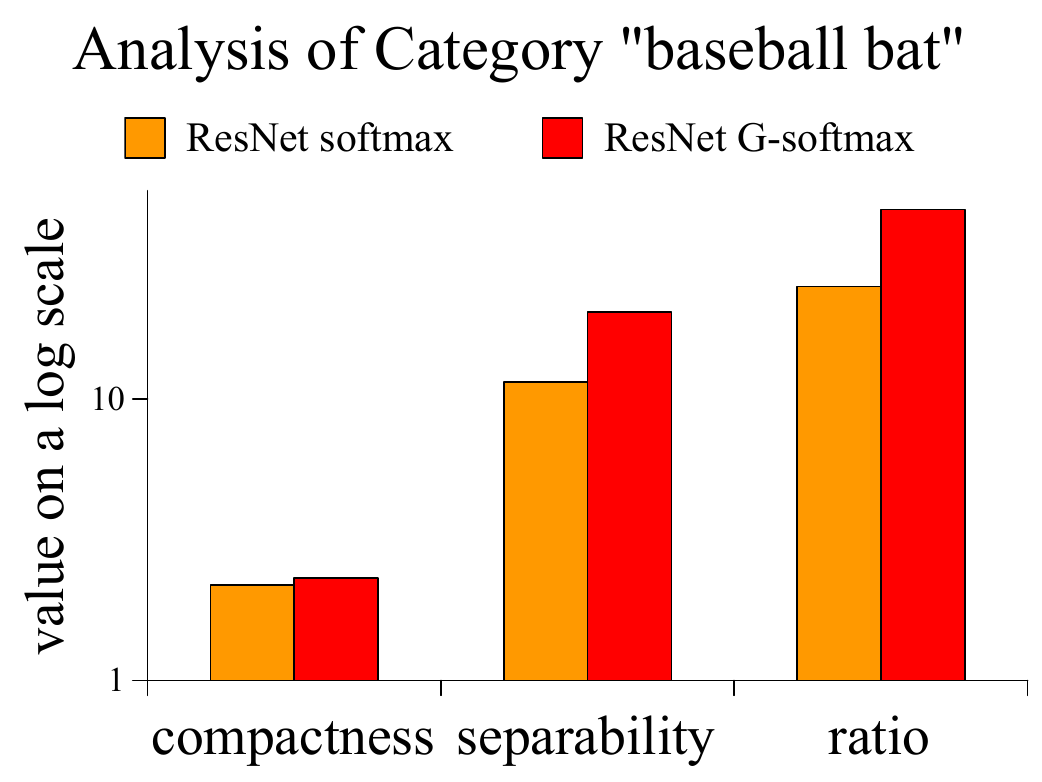} \\
	\end{tabular}
	\vspace{-2ex}
	\caption
	{
		\small
		Feature analysis of the softmax function and the $\mathcal{G}$-softmax function on MSCOCO. The first row is the results of category ``cow'' while the second row is the results of category ``baseball bat''. For each row, from left to right, the first and second column is $x$ vs. $p$ plot. The third column is the Gaussian distributions of $x$ in the first and second column. The last column is the corresponding compactness, separability and ratio. These plots show that the $\mathcal{G}$-softmax function improves both the compactness (the positive curve in the third column becomes taller and narrower) and the separability (the positive curve and the negative curve are farer away).
	}
	\label{fig:anl_mscoco}
	\vspace{-1ex}
\end{figure*}

\noindent\textbf{Notations}. We denote a model with the $\mathcal{G}$-softmax function as \textit{model $\mathcal{G}$-softmax}, \eg ResNet-101 $\mathcal{G}$-softmax. To simplify notations, we omit \text{softmax} following the model name because we assume that the models work with the softmax function by default. For example, \text{ResNet} implies that the ResNet model works with the softmax function. In Table \ref{tbl:perf} and Figure \ref{fig:scatter}, \ref{fig:gaussians}, \ref{fig:anl_mscoco}, \ref{fig:cifar10}, and \ref{fig:cifar100}, RSN, DSN, and WRN stand for ResNet, DenseNet, and wide ResNet, respectively.

\noindent\textbf{Evaluations on CIFAR}. The performances of the softmax function and the $\mathcal{G}$-softmax function are listed in Table~\ref{tbl:perf_train} in terms of top-1 error rate. For the convenient purpose, DenseNet and wide ResNet are denoted as DSN and WRN, respectively. The proposed $\mathcal{G}$-softmax function outperforms softmax and L-softmax function over all evaluated scenarios.

On CIFAR-10, VGG with the $\mathcal{G}$-softmax function achieves $5.54\%$ error rate while the error rates of the softmax and L-softmax function are $5.69\%$ and $7.79\%$, respectively. Consistently, VGG with the $\mathcal{G}$-softmax function achieves the similar improvement on CIFAR-100. DenseNet reports their best error rate on CIFAR-10 and CIFAR-100 with 190 convolutional layers and 40 growth rate (denoted as DSN-BC-190-40)~\cite{Huang_CVPR_2017}. However, DenseNet with this configuration consumes huge graphics memory due to the large depth number, which would occupy about 30 GB of graphics memory to process a batch of 10 images on 3 GPUs. Therefore, we adopt a moderate setting, \ie DSN-100-24, in our experiments to process as large batch size as possible, \ie 50 on CIFAR-10 and 32 on CIFAR-100. Under this configuration, the $\mathcal{G}$-softmax function achieves $3.70\%$ error rate, which is better than the error rate $3.77\%$ of the softmax function and the error rate $4.84\%$ of the L-softmax function, on CIFAR-10. Also, the error rate of the $\mathcal{G}$-softmax function is decreased to $18.89\%$ compared to the error rate $19.25\%$ of the softmax function and the error rate of $23.22\%$ of the L-softmax function on CIFAR-100. In wide ResNet experiments, the baseline consistently achieves better performances than the baseline of DenseNet on both CIFAR-10 and CIFAR-100, where the $\mathcal{G}$-softmax function further improves the performances to achieve error rate $3.36\%$ on CIFAR-10 and $17.41\%$ on CIFAR-100. As shown in Table~\ref{tbl:perf_train}, although the structures of the three model are distinct to each other, the $\mathcal{G}$-softmax function generalize to these models and improve the respective performances. Applying malleable learning rates with wide ResNet $\mathcal{G}$-softmax can further improve the performances, \ie $3.14\%$ on CIFAR-10 and $17.04\%$ on CIFAR-100.

\noindent\textbf{Evaluations on Tiny ImageNet}.
Table \ref{tbl:timg_perf} reports the error rates of softmax, L-softmax, and the proposed $\mathcal{G}$-softmax function on Tiny ImageNet. We present the error rates of ResNet with input image size $64\times64$ and $224\times224$, where $224\times224$ is used in the setting of training on ImageNet and the training of the initialized ResNet fed with this image size leads to a lower error rate $18.36\%$. The proposed $\mathcal{G}$-softmax function with various $(\mu, \sigma)$ leads to overall lower error rates than the softmax and L-softmax function. In particular, $(\mu=-0.05, \sigma=1)$ achieves the lowest error rate $16.86\%$.

\noindent\textbf{Evaluations on MS~COCO}.
As shown in Table \ref{tbl:mscoco_perf}, ResNet-101 $\mathcal{G}$-softmax with an initialization of Gaussian distributions $(-0.1,1)$ for $(\mu, \sigma)$ achieves the best performance over three metrics (\ie C-F1, O-F1, and mAP). The proposed $\mathcal{G}$-softmax functions are initialized in two straightforward ways. One is to set $(\mu, \sigma)$ to the standard Gaussian distribution parameter $(0,1)$, while the other one is to empirically compute $(\mu, \sigma)$ from the data. Both approaches achieve  better mAPs ($80.8\%$ and $81.0\%$) than the state-of-the-art model \cite{Durand_CVPR_2017} ($80.7\%$). To comprehensively understand the effects of $\mu,\sigma$, we initialize them with other values, \ie $(\pm0, 0.5)$, $(\pm0, 5)$, and $(\pm0.1, 1)$, By comparing with the performance of ResNet-101 $\mathcal{G}$-softmax with $(0,1)$, we can see the respective influences of $\mu,\sigma$. Overall, the four initializations leads to better performances than the initialization of $(0,1)$ and the initialization of $(-0.1,1)$ yields the best performance over C-F1, O-F1, and mAP. An observation on $\mu$ is that smaller $\sigma$ leads to higher precision but lower recall. For example, the O-P of $\sigma=0.5$ is $83.5\%$ whereas the one of $\sigma=5$ is $81.3\%$. Nevertheless, the O-R of $\sigma=0.5$ is $72.9\%$ whereas the one of $\sigma=5$ is $74.7\%$. 
According to metrics (\ref{eqn:mc_metrics}), we can infer that small $\sigma$ yields less $N_{i}^{c}$ and $N_{i}^{p}$ than large $\sigma$. The change in $N_{i}^{c}$ is relatively smaller than the one in $N_{i}^{p}$ and these effects of decreasing $\sigma$ lead to higher precision but lower recall.

\noindent\textbf{Evaluations on NUS-WIDE}.
The experimental results of NUS-WIDE are consistent with the experimental results of MS COCO, as shown in Table \ref{tbl:nus_perf}. The proposed $\mathcal{G}$-softmax function overall outperforms the softmax function over all metrics. Specifically, the setting $(\mu=0.05, \sigma=1)$ achieves the best mAP $60.4\%$.

%% file: depd/analysis.tex
\section{Analysis}

In this section, we discuss the influence of the proposed $\mathcal{G}$-softmax function on prediction by presenting a visual comparison with the softmax and L-softmax function. Then, we further quantify the influences caused by the softmax, L-softmax, and the proposed $\mathcal{G}$-softmax function in terms of intra-class compactness and inter-class separability. 
Moreover, the analysis of significance of the AP differences between the softmax function and the proposed $\mathcal{G}$-softmax function on MS COCO and NUS-WIDE is provided.
Last but not least, we analyze the correlations between compactness (separability and ratio) and AP on MS COCO and NUS-WIDE.

\noindent\textbf{Influence of the $\mathcal{G}$-softmax function on ConvNets}.
In the literature, there are many works \cite{Zeiler_ECCV_2014, Mahendran_CVPR_2015, Lenc_CVPR_2015} that analyze ConvNets using visualization. In this work, our hypothesis is related to the distributions of the activations of deep layers. Therefore, we analyze the proposed $\mathcal{G}$-softmax function from the aspect of the mapping between $x$ and $p$. Given images with a certain label $c$ out of $m$ labels, ConvNets would generate the final feature $x\in \mathbb{R}^{m}$ preceding to the process of the softmax function. Each $x_{i}$ in $x$ represents the corresponding confidence for the predicted label $i$. By the idea of winner-takes-all in the softmax function, the corresponding label $i$ that has the highest value $p$ of the softmax function would be marked as the prediction. We hope that the predicted label is the ground truth label, \ie $i=c$, and name $j,j\ne c$ imposter labels. Ideally, the imposter feature $x_{j}$ is expected to be lower and far away from the ground truth feature $x_{i}$ so as to enlarge the probability of correct prediction.

\begin{figure*}[!t]
	\centering
	\begin{tabular}{cccc}
		\includegraphics[width=0.228\textwidth]{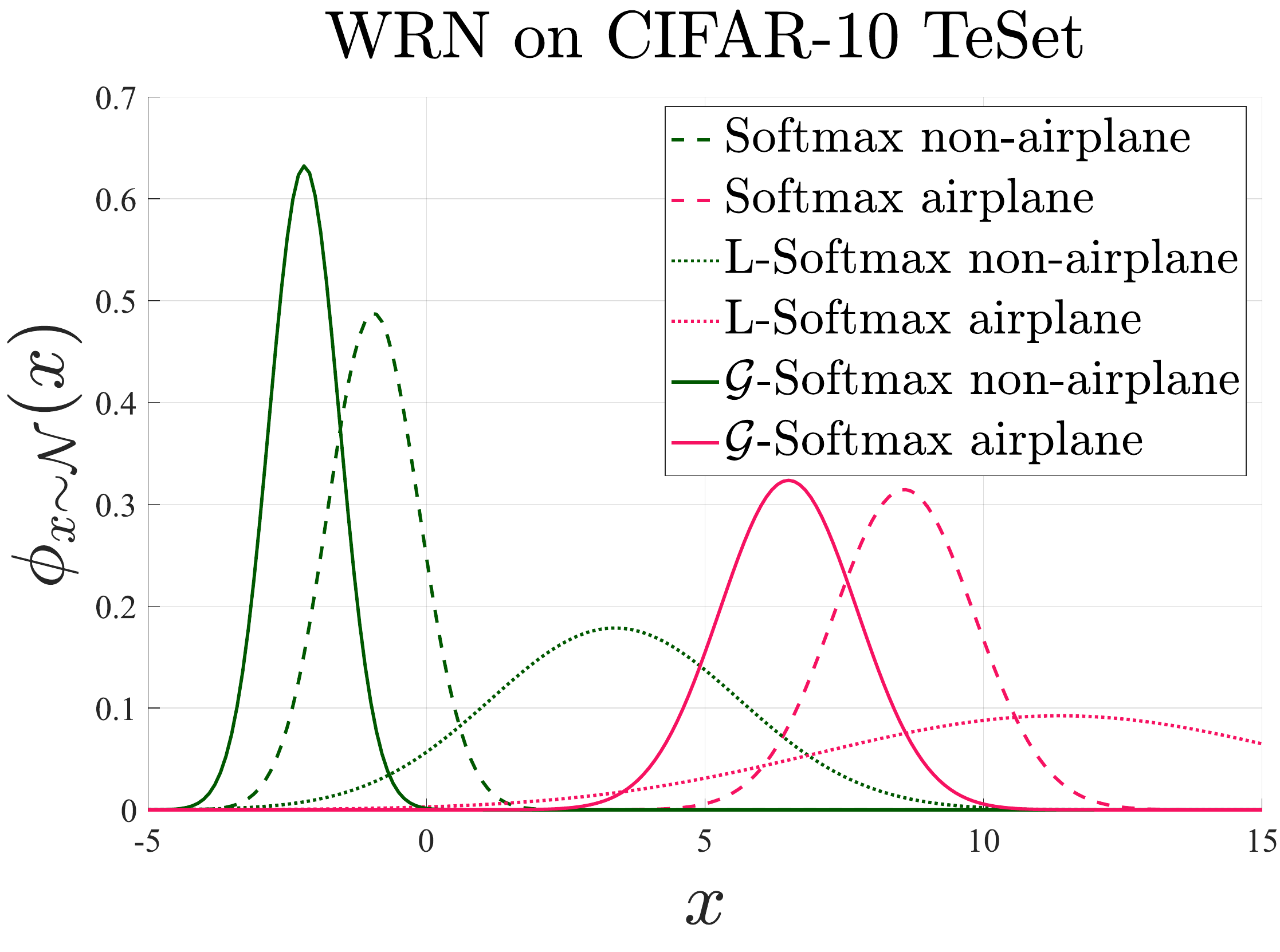} 			 &  \includegraphics[width=0.228\textwidth]{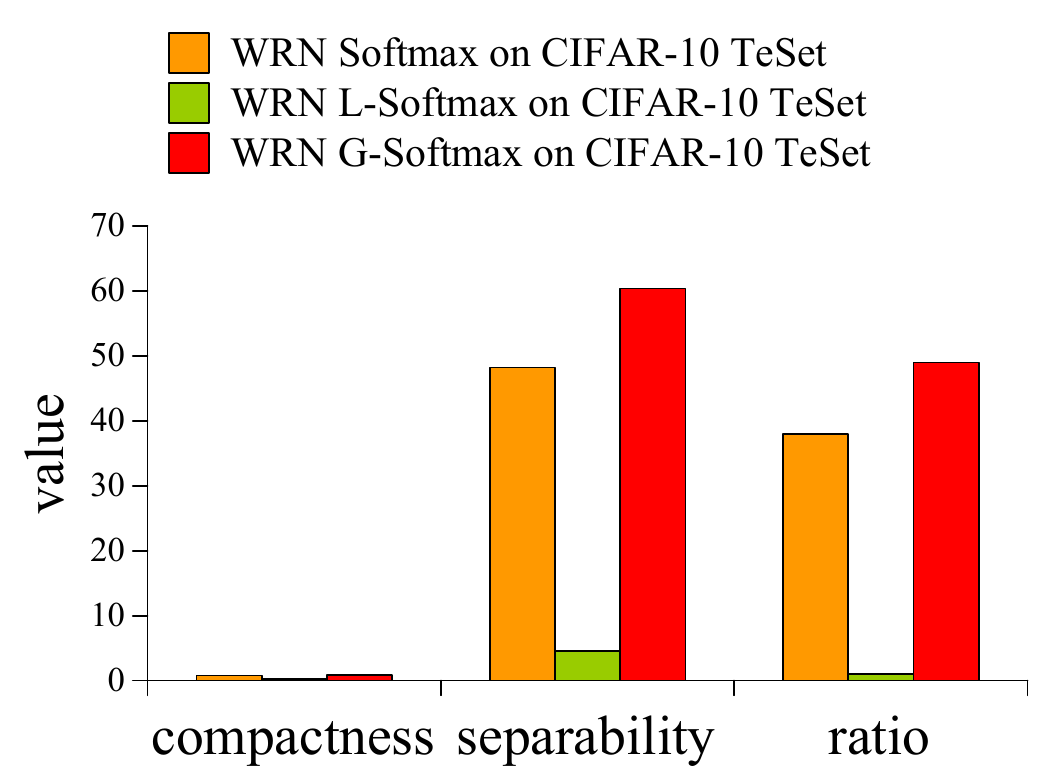} &
		\includegraphics[width=0.228\textwidth]{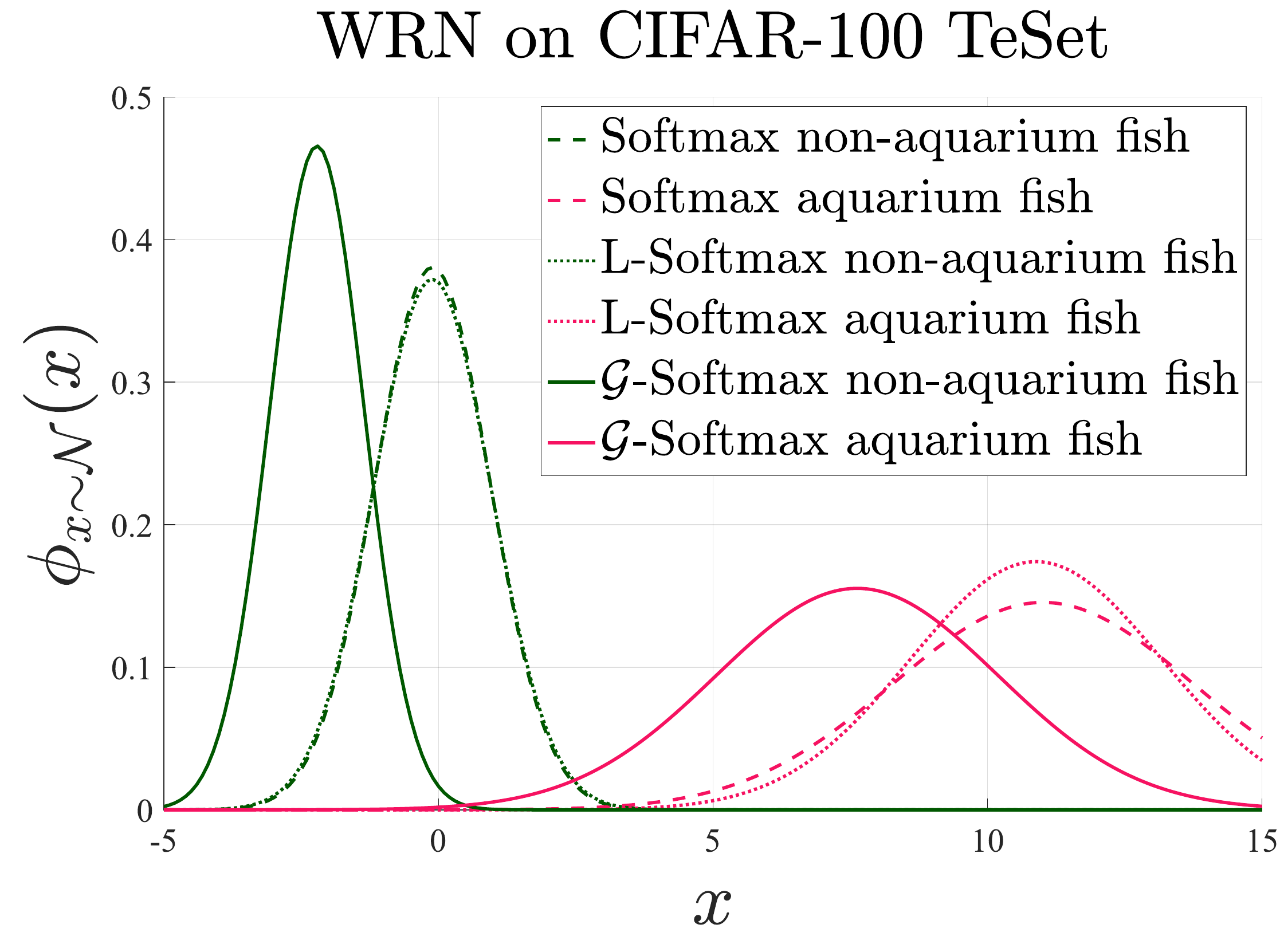} 			 &  \includegraphics[width=0.228\textwidth]{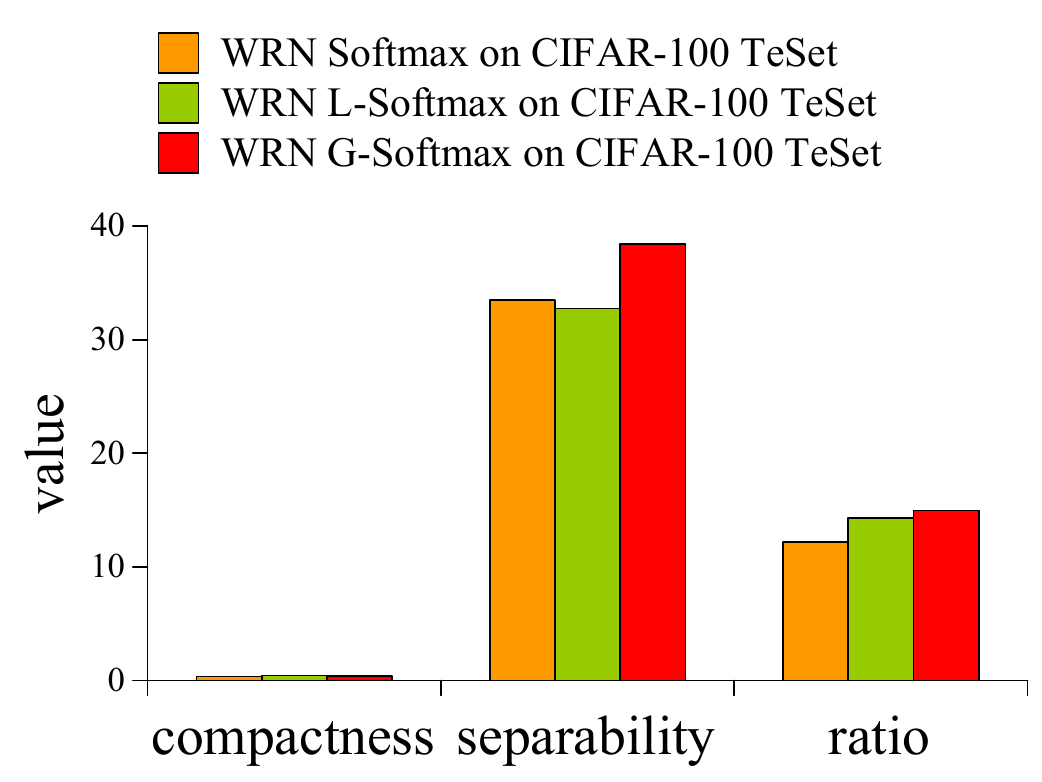} \\
	\end{tabular}
	\vspace{-2ex}
	\caption
	{
		\small
		Gaussian distributions of $x$ and corresponding compactness, separability, and ratio on the test set of Figure~\ref{fig:scatter}. In the CIFAR-10 experiments, given all the testing images with respect to ground truth class ``airplane'', given based on $x_{1}$, we compute the empirical $\mu_{1}$ and $\sigma_{1}$ so that the compactness, separability, and ratio can be computed. Similar procedure is conducted in the CIFAR-100 experiments. It can be seen that the ratios of the proposed $\mathcal{G}$-softmax function are overall better than the ones of softmax and L-softmax function.
	}
	\label{fig:gaussians}
\end{figure*}

To investigate the influence of the trained $\mathcal{G}$-softmax function on the training set and test set, we inspect the relationship between features $x$ and predictions $p$ on CIFAR-10 and CIFAR-100, as shown in Figure \ref{fig:scatter}. To remove unnecessary interference from patterns of other classes, we fix the prediction of a subset of the training set and the test set of CIFAR-10 from a single class. For example, given all images with the ground truth class label ``airplane'', the ConvNet would generate the deep features {\small $x \in \mathbb{R}^{m}$}, $m=10$ in CIFAR-10, and pass them to the predictor for computing the predictions $p$. Note that here we denote $x_{1}$ as the feature of the class ``airplane'' and all $x_{j},(j\ne 1)$ are considered the features w.r.t. ``non-airplane''. Similarly, we also plot the scattered points w.r.t. the images with label ``aquarium fish'' on CIFAR-100.

As shown in Figure \ref{fig:scatter}, the range of $x$ of the proposed $\mathcal{G}$-softmax function is different from the range of $x$ of the softmax and L-softmax function. The most of imposter features $x_{j}$ of the proposed $\mathcal{G}$-softmax function are distributed in the range $[-5,0]$, whereas $x_{j}$ of the softmax and L-softmax function spreads out.
In the test set of CIFAR-10, the range of $x_{c}$ of the proposed $\mathcal{G}$-softmax function approximately spans from 0 to 9, whereas the range of the softmax function is $[0, 11]$ and the range of the L-softmax function is $[0, 24]$. 
In the test set of CIFAR-100, the range of $x_{c}$ of the proposed $\mathcal{G}$-softmax function approximately spans from 0 to 11, whereas the range of the softmax function is $[0, 15]$ and the range of the L-softmax function is $[0, 14]$. 

Figure~\ref{fig:anl_mscoco} w.r.t. two categories on MS~COCO shows consistent pattern. In category ``cow'' and ``baseball bat'', the positive features of ResNet-101 $\mathcal{G}$-softmax, i.e., the features related to ``cow'' and ``baseball bat'', are closer to each other than the ones of ResNet-101 with the softmax function.

To quantitatively understand the distributions of the scattered points in Figure \ref{fig:scatter}, we empirically compute $\mu$ and $\sigma$ of the points w.r.t. the softmax function, the L-softmax function, and the proposed $\mathcal{G}$-softmax function. With theses distribution parameters, we further compute the compactness, separability, and ratio, as shown in Figure \ref{fig:gaussians}.

\begin{figure*}[t!]
	\centering
	\includegraphics[width=0.8\textwidth]{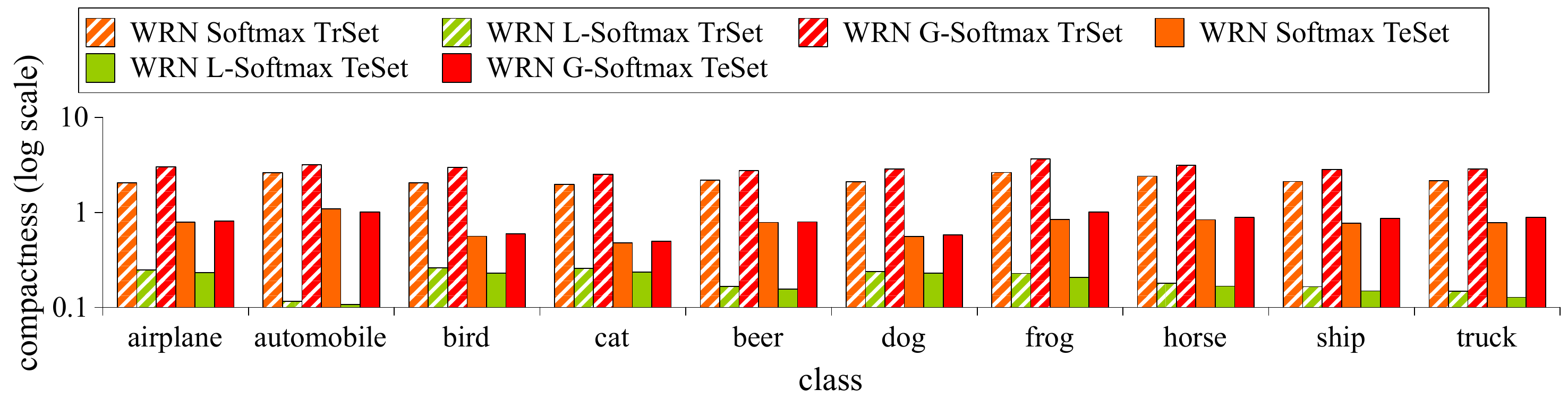}
	\includegraphics[width=0.8\textwidth]{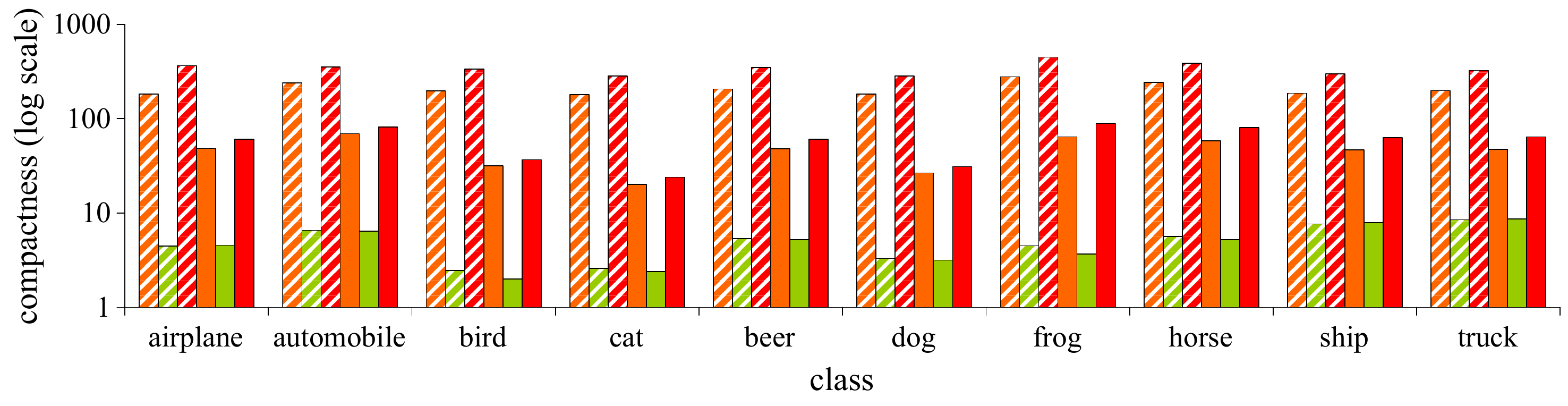}
	\includegraphics[width=0.8\textwidth]{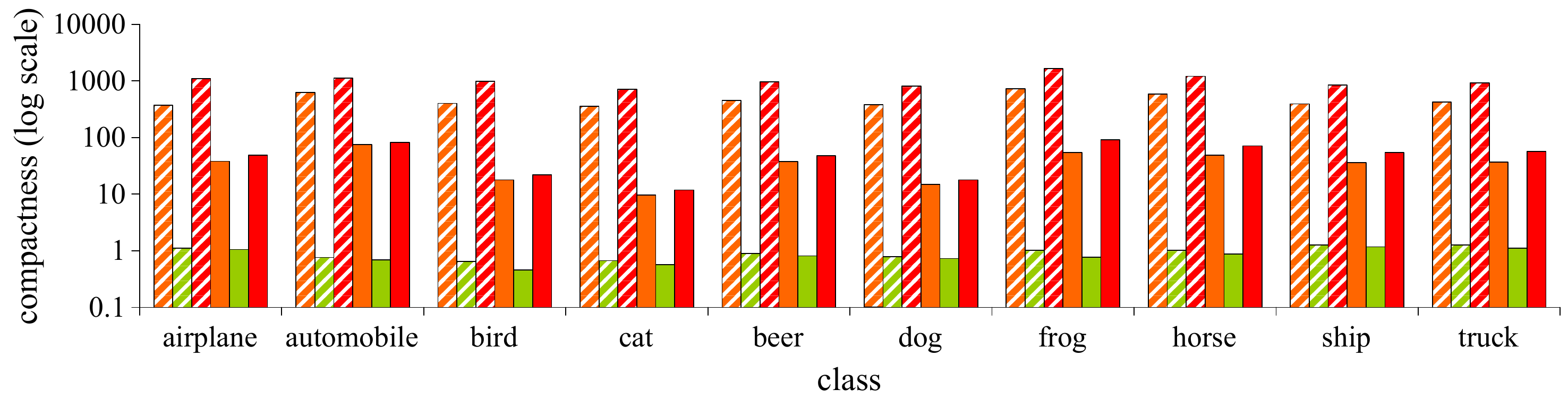}
	\vspace{-1ex}
	\caption{Analysis on CIFAR-10 test set in terms of compactness, separability, and separability-$\sigma$ ratio over each class with wide ResNet. 
	We can see that the proposed $\mathcal{G}$-softmax function improves compactness, separability, and ratio on both training and test set in most categories.}
	As discussed in Section \ref{sec:method}, the compactness is defined as the reciprocal of $\sigma$.
	\label{fig:cifar10}
	\vspace{-1ex}
\end{figure*}

The proposed $\mathcal{G}$-softmax function influences the kurtosis of the Gaussian distributions of $x$ of class `airplane' (CIFAR-10) or `aquarium fish' (CIFAR-100), comparing to the softmax function. 
In other words, the curves of the distributions w.r.t. the proposed $\mathcal{G}$-softmax function are narrower and taller than the ones w.r.t. the softmax function on both CIFAR-10 and CIFAR-100. In particular, the distributions w.r.t. the L-softmax function yields a flatter and wider curves than  the softmax function and the proposed $\mathcal{G}$-softmax function on CIFAR-10 and CIFAR-100. With the distribution parameters, the intra-class compactness, inter-class separability, and separability-$\sigma$ ratio can be computed and visualized in the bar plots in Figure \ref{fig:gaussians}. Overall, the proposed $\mathcal{G}$-softmax function achieves better intra-class compactness, inter-class separability, and separability-$\sigma$ ratio than the softmax function and the L-softmax function on both CIFAR-10 and CIFAR-100.

\begin{figure*}[t!]
	\centering
	\includegraphics[width=0.8\textwidth]{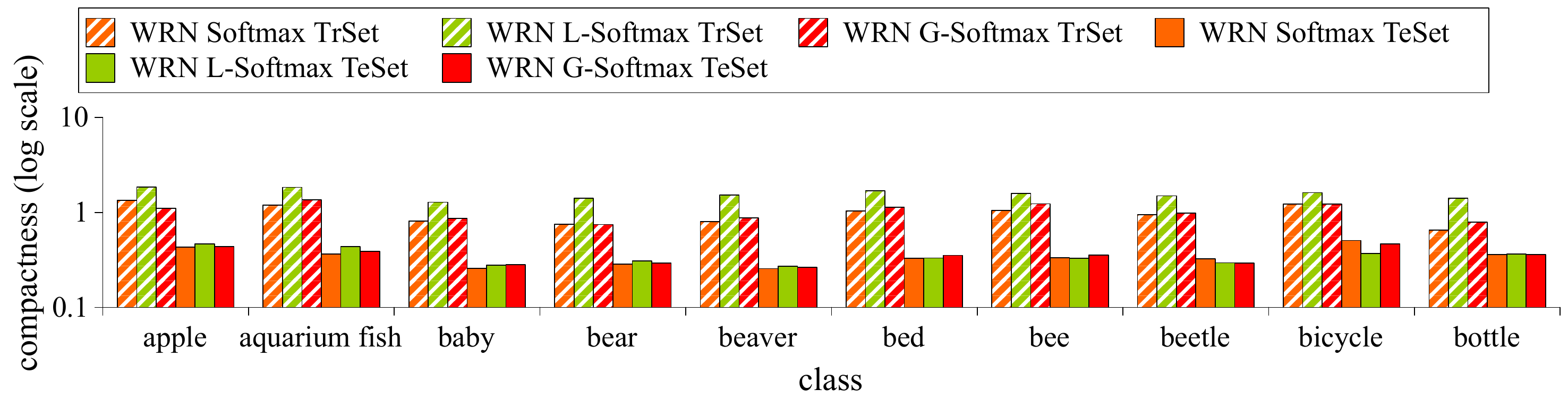}
	\includegraphics[width=0.8\textwidth]{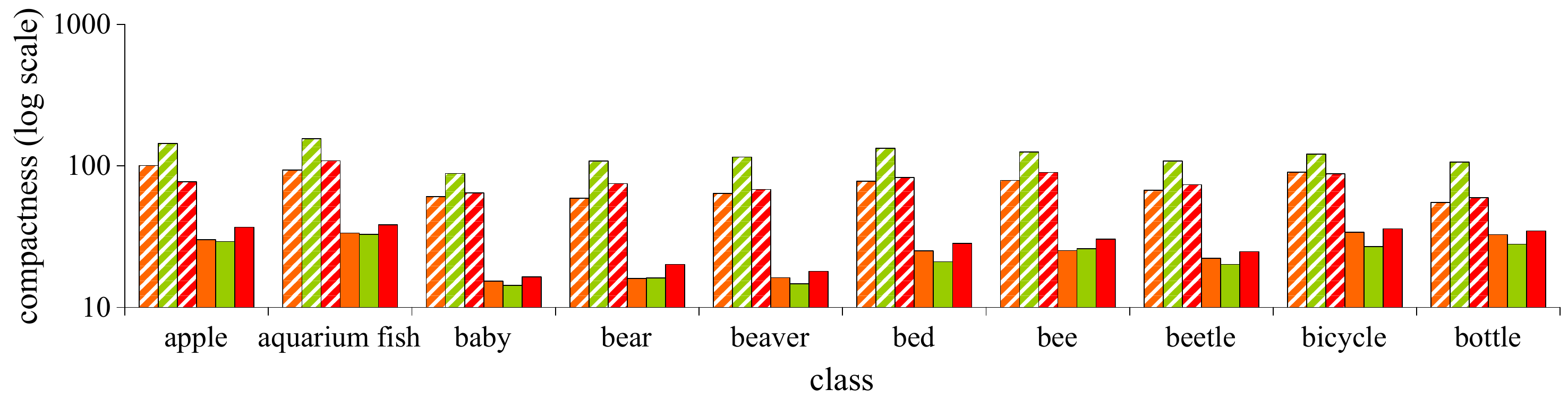}
	\includegraphics[width=0.8\textwidth]{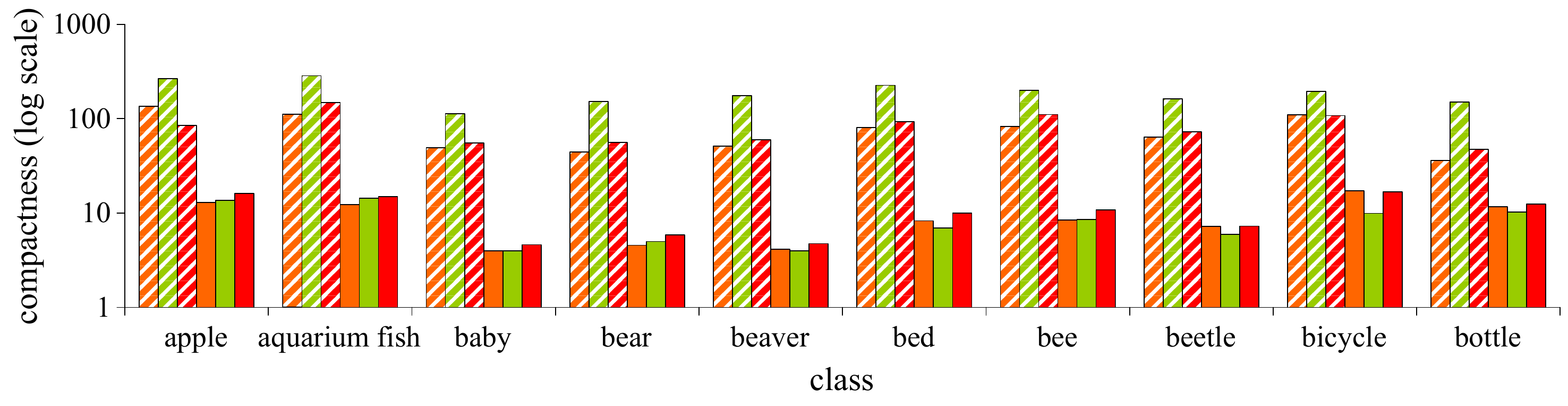}
	\vspace{-1ex}
	\caption{Analysis on CIFAR-100 test set in terms of compactness, separability, and separability-$\sigma$ ratio with wide ResNet. For clarity, we present the analyses on the first 10 classes.
	Although L-softmax achieves better scores over compactness, separability, and ratio on the training set, it has much lower scores on the test set. This implies that it overfits the training set.
	In contrast, we can see that the proposed $\mathcal{G}$-softmax function improves compactness, separability, and ratio on both training and test set in most categories.
	}
	\label{fig:cifar100}
	\vspace{-1ex}
\end{figure*}

Figure \ref{fig:cifar10} shows more comprehensive analysis intra-class compactness, inter-class separability, and separability-$\sigma$ ratio for each class on CIFAR-10. We can see that the proposed $\mathcal{G}$-softmax function improves intra-class compactness, inter-class separability, and separability-$\sigma$ ratio in most of classes over the softmax function and the L-softmax function. Due to the limitation of space, we do the similar analysis on the first 10 classes on CIFAR-100, as shown in Figure \ref{fig:cifar100}. In contrast to Figure \ref{fig:cifar10}, where the L-softmax function yields the lowest intra-class compactness, inter-class separability, and separability-$\sigma$ ratio on both the training and test set of CIFAR-10, the L-softmax function yields the highest intra-class compactness, inter-class separability, and separability-$\sigma$ ratio in most of classes on the training set but still yields the lowest intra-class compactness, inter-class separability, and separability-$\sigma$ ratio in most of classes on the test set. This implies that it may overfit the training data. Again, the proposed $\mathcal{G}$-softmax function consistently yields better intra-class compactness, inter-class separability, and separability-$\sigma$ ratio in most of classes.

\begin{figure}[t!]
	\centering
	\includegraphics[width=.3\textwidth]{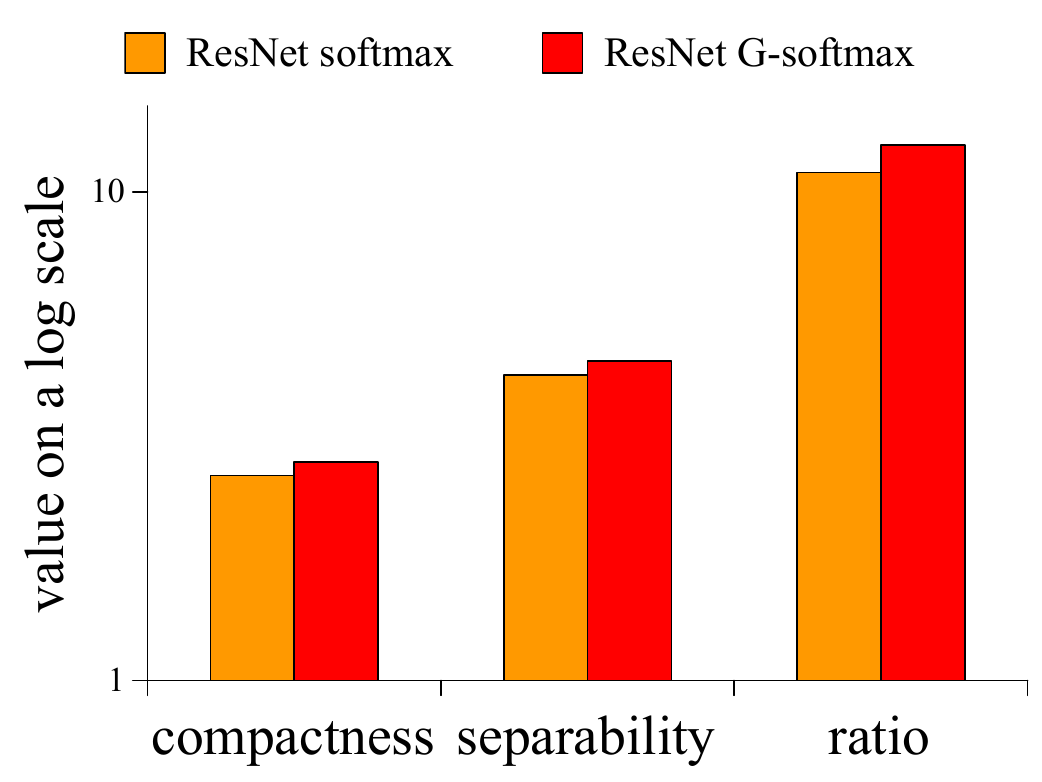}
	\caption{
		The average compactness, separability, and ratio over all 80 categories on MS~COCO validation set. The $\mathcal{G}$-softmax function give rise to improvements on all metrics.
	}
	\label{fig:coco_all}
\end{figure}

We also analyze intra-class compactness, inter-class separability, and separability-$\sigma$ ratio for multi-label classification on MS COCO. 
The experimental results of MS COCO show a consistent pattern with the ones of CIFAR. For example, the $x$ vs. $p$ plots of category ``baseball bat'' in Figure \ref{fig:anl_mscoco} show that $x$ of ResNet $\mathcal{G}$-softmax are more compact than $x$ of ResNet. Consistently, the Gaussian distribution of ResNet $\mathcal{G}$-softmax w.r.t. the positive $x$ is taller and narrower than the one of ResNet. The compactness of ResNet w.r.t. class ``baseball bat'' is 2.1, while the one of ResNet $\mathcal{G}$-softmax is 2.3. Figure \ref{fig:coco_all} shows the average compactnesses of ResNet and ResNet $\mathcal{G}$-softmax over all 80 categories on MS~COCO validation set. The average compactness of ResNet is 2.6, while the one of ResNet $\mathcal{G}$-softmax is 2.8. The separability of the proposed $\mathcal{G}$-softmax function between category ``non-cow'' and ``cow'' is 4.3, which is significantly greater than 1.8 (\ie the separability of the softmax function). The average separability over all 80 categories on MS~COCO is shown in Figure \ref{fig:coco_all}. The average separability (4.5) of the proposed $\mathcal{G}$-softmax function is greater than the average separability (4.2) of the softmax function. Similar to intra-class compactness and inter-class separability, the average ratio of the proposed $\mathcal{G}$-softmax function is higher than the one of the softmax function.

\begin{table}[t!]
	\centering
	\vspace{-1ex}
	\caption{
		The analysis of the significance of the prediction differences between the softmax function and various $\mathcal{G}$-softmax functions in Table \ref{tbl:mscoco_perf} and \ref{tbl:nus_perf}. The significance of APs w.r.t. the softmax function and the proposed $\mathcal{G}$-softmax function is computed by paired-sample t-test. The resulting $p$-val $ \in [0,1]$ reported in the table is the probability assuming the null hypothesis were true. If $p$-val is equal to or less than 0.05, it implies that there is a significant difference between the softmax function and the proposed $\mathcal{G}$-softmax function in compactness (separability or ratio). In the experiments on MS COCO, the difference of APs between the softmax function and the proposed $\mathcal{G}$-softmax function with $\mu=0,\ \sigma=0.5$ is statistically significant ($p\text{-val} < 0.05$). In the experiments on NUS-WIDE, besides the proposed $\mathcal{G}$-softmax function with $\mu=-0.05,\ \sigma=1$, the differences of APs between the softmax function and the proposed $\mathcal{G}$-softmax functions are statistically significant.
	}
	\vspace{-1ex}
	\scalebox{0.9}{
	\begin{tabular}{lc}
		\toprule
		& MS COCO \\
		\cmidrule(lr){2-2} 
		$\mathcal{G}$-softmax(0,1) & 0.9060  \\
		$\mathcal{G}$-softmax(0,0.5) & \textbf{0.0359} \\
		$\mathcal{G}$-softmax(0,5) & 0.0548 \\
		$\mathcal{G}$-softmax(-0.1,1) & 0.0764 \\
		$\mathcal{G}$-softmax(0.1,1) & 0.3160 \\
		\bottomrule
	\end{tabular}} \hfill
	\scalebox{0.9}{
	\begin{tabular}{lc}
		\toprule
		& NUS-WIDE \\
		\cmidrule(lr){2-2} 
		$\mathcal{G}$-softmax(0,1) & 0.0049  \\
		$\mathcal{G}$-softmax(0,2) & \textbf{0.0001} \\
		$\mathcal{G}$-softmax(0,3) & 0.0098 \\
		$\mathcal{G}$-softmax(-0.05,1) & 0.6773 \\
		$\mathcal{G}$-softmax(0.05,1) & 0.0131 \\
		\bottomrule
	\end{tabular}
	}
	\vspace{-2ex}
	\label{tbl:significance}
\end{table}

\noindent\textbf{Significance of Difference between softmax and $\mathcal{G}$-softmax}. 
As aforementioned discussion about the influence of the proposed $\mathcal{G}$-softmax function, we further quantify the difference of prediction performance caused by the influence. Specifically, we study the difference of average precision between the softmax function and the proposed $\mathcal{G}$-softmax function on MS COCO and NUS-WIDE, which are richer in visual content and visual semantics than CIFAR and Tiny ImageNet. First, APs of the softmax function and the proposed $\mathcal{G}$-softmax function w.r.t. each class are computed, respectively. Particularly, the proposed $\mathcal{G}$-softmax functions with each pair of $\mu$ and $\sigma$ in Table \ref{tbl:mscoco_perf} and \ref{tbl:nus_perf} are used for analysis. With APs of the softmax function and APs of the proposed $\mathcal{G}$-softmax function with a specific $\mu$ and $\sigma$, paired-sample t-test will be used to compute p value denoted as $p$-val, which indicates the probability assuming the null hypothesis were true. When $p\text{-val} \le 0.05$, this implies that the pair of two series of APs are significantly different. Table \ref{tbl:significance} shows such an analysis on MS COCO and NUS-WIDE, respectively. We can see that $p$-val of the softmax function and the proposed $\mathcal{G}$-softmax function with $\mu=0,\ \sigma=0.5$ is less than 0.05 in the experiments on MS COCO. This implies the resulting APs of the proposed $\mathcal{G}$-softmax function are significantly different from the ones of the softmax function. In contrast, in the experiments on NUS-WIDE, the proposed $\mathcal{G}$-softmax functions in Table \ref{tbl:nus_perf} are significantly different from the softmax function in terms of APs, other than the proposed $\mathcal{G}$-softmax function with $\mu=-0.05,\ \sigma=1$.

\noindent\textbf{Correlations between Compactness/Separability/ratio and APs}.
In this work, we study intra-class compactness and inter-class separability for each class in the datasets. A question comes up, that is, how are intra-class compactness and inter-class separability correlated to APs in the proposed $\mathcal{G}$-softmax function? Note that intra-class compactness and inter-class separability may not be influential when the values of them are low. Hence, we only inspect the classes with best average intra-class compactness, inter-class separability, or separability-$\sigma$ ratio across various $\mathcal{G}$-softmax functions. On one hand, we have intra-class compactnesses (inter-class separabilities or separability-$\sigma$ ratios) of these classes w.r.t. each $\mathcal{G}$-softmax functions in Table \ref{tbl:mscoco_perf} and \ref{tbl:nus_perf}. On the other hand, we have the APs yielded by each $\mathcal{G}$-softmax functions in Table \ref{tbl:mscoco_perf} and \ref{tbl:nus_perf}. With the compactness/separabilities/ratios and corresponding APs of a certain class yielded by various $\mathcal{G}$-softmax functions, we use Pearson correlation method to quantify the correlation between the three factors and AP and report Pearson correlation coefficients and corresponding $p$-vals in Table \ref{tbl:corr}. We can observe that overall intra-class compactness, inter-class separability, or separability-$\sigma$ ratio are linearly correlated to AP to a significance level of 0.05. This implies improvement of intra-class compactness and inter-class separability will leads to improvement of AP.

\begin{table}[!t]
	\centering
	\caption{
	Correlations between compactness (separability and ratio) and AP across various proposed $\mathcal{G}$-softmax functions in Table \ref{tbl:mscoco_perf} and \ref{tbl:nus_perf} on MS COCO and NUS-WIDE. As each class has its own underlying distribution, we first find the class with best average compactness, separability, or ratio. Then, the compactnesses (separability or ratio) of this class across various $\mathcal{G}$-softmax functions in Table \ref{tbl:mscoco_perf} and \ref{tbl:nus_perf} are used to compute the Pearson correlation with the corresponding APs of various $\mathcal{G}$-softmax functions. The Pearson correlation coefficient and the corresponding value are reported as $(\rho, p\text{-val})$ in the table. $\rho$ is in $[-1,1]$. When $\rho=1$, it indicates that compactness (separability or ratio) is perfectly linearly correlated to AP. We can see that compactness (separability or ratio) of these classes are linearly correlated to APs to a significance level of 0.05.
	}
	\vspace{-1ex}
	\begin{tabular}{lcc}
		\toprule
		& MS COCO & NUS-WIDE \\
		\cmidrule(lr){2-2} \cmidrule(lr){3-3}
		Correlation(compactness, AP) & (0.9472, 0.0144) & (0.9635, 0.0083)  \\
		Correlation(separability, AP) & (0.9791, 0.0036) & (0.9045, 0.0349) \\
		Correlation(ratio, AP) & (0.9636, 0.0083) & (0.9702, 0.0062) \\
		\bottomrule 
		\label{tbl:corr}
	\end{tabular}
	\vspace{-2ex}
\end{table}

%% file: depd/conclusion.tex
\section{Conclusion}

In this work, we propose a Gaussian-based softmax function, namely $\mathcal{G}$-softmax, which uses cumulative probability function to improve features' intra-class compactness and inter-class separability. The proposed $\mathcal{G}$-softmax function is simple to implement and can easily replace the softmax function. For evaluation purposes, classification datasets (\ie CIFAR-10, CIFAR-100 and Tiny ImageNet) and on multi-label classification datasets (\ie MS COCO and NUS-WIDE) are used in this work. The experimental results show that the proposed $\mathcal{G}$-softmax function improves the state-of-the-art ConvNet models. Moreover, in our analysis, it is observed that high intra-class compactness and inter-class separability are linearly correlated to average precision on MS COCO and NUS-WIDE.